\documentclass[10pt,twocolumn,letterpaper]{article}

\usepackage{cvpr}
\usepackage{times}
\usepackage{epsfig}
\usepackage{graphicx}
\usepackage{amsmath}
\usepackage{amssymb}

\usepackage{subfigure}
\usepackage{comment}
\usepackage{epstopdf}
\usepackage{multirow,bigstrut}
\usepackage{gensymb}

\usepackage[pagebackref=true,breaklinks=true,letterpaper=true,colorlinks,bookmarks=false]{hyperref}

 \cvprfinalcopy % *** Uncomment this line for the final submission

\def\cvprPaperID{2330} % *** Enter the CVPR Paper ID here

\ifcvprfinal\fi
\begin{document}

%%%%%%%%% TITLE
\title{Optimal Radiometric Calibration for\\ Camera-Display Communication}

 \author{ Wenjia Yuan \and Eric Wengrowski \and Kristin J. Dana \and Ashwin Ashok \and Marco Gruteser  \and Narayan Mandayam \\
 Department of Electrical and Computer Engineering, and \\
              WINLAB (Wireless Information Network Laboratory)
              Rutgers University \\
              Tel.: +732-445-5253\\
              Contact Author: {\tt\small kdana@ece.rutgers.edu} \\
}

\maketitle
%\thispagestyle{empty}

%%%%%%%%% ABSTRACT
\begin{abstract}
We present a novel method for communicating between a camera and display by embedding and recovering hidden and dynamic information within a displayed image.  
A handheld camera pointed at the display can receive not only the display image, but also the underlying message. 
These active scenes are fundamentally different from traditional  passive scenes like QR codes because image formation is based on display emittance, not surface reflectance.
Detecting  and decoding the message requires careful photometric modeling for computational message recovery.  
Unlike standard watermarking and steganography methods that lie outside the domain of computer vision, our message recovery algorithm uses illumination to optically communicate hidden messages in real world scenes. 
The key innovation of our approach is an algorithm that performs simultaneous radiometric calibration and message recovery in one convex optimization problem.
By modeling the photometry of the system using a camera-display transfer function (CDTF), we derive a physics-based kernel function for support vector machine classification.  
We demonstrate that our method of {\it optimal online radiometric calibration (OORC)} leads to an efficient and robust algorithm for computational messaging between nine commercial cameras and displays.  

\end{abstract}

%%%%%%%%% BODY TEXT

%%%%%%%%%% INTRODUCTION %%%%%%%%%%%
\section{Introduction}\label{sec:intro}

% Background, Objectives, brief summary of methods

%
\begin{figure}[t]
	\begin{center}
		\includegraphics[width=3.5in]{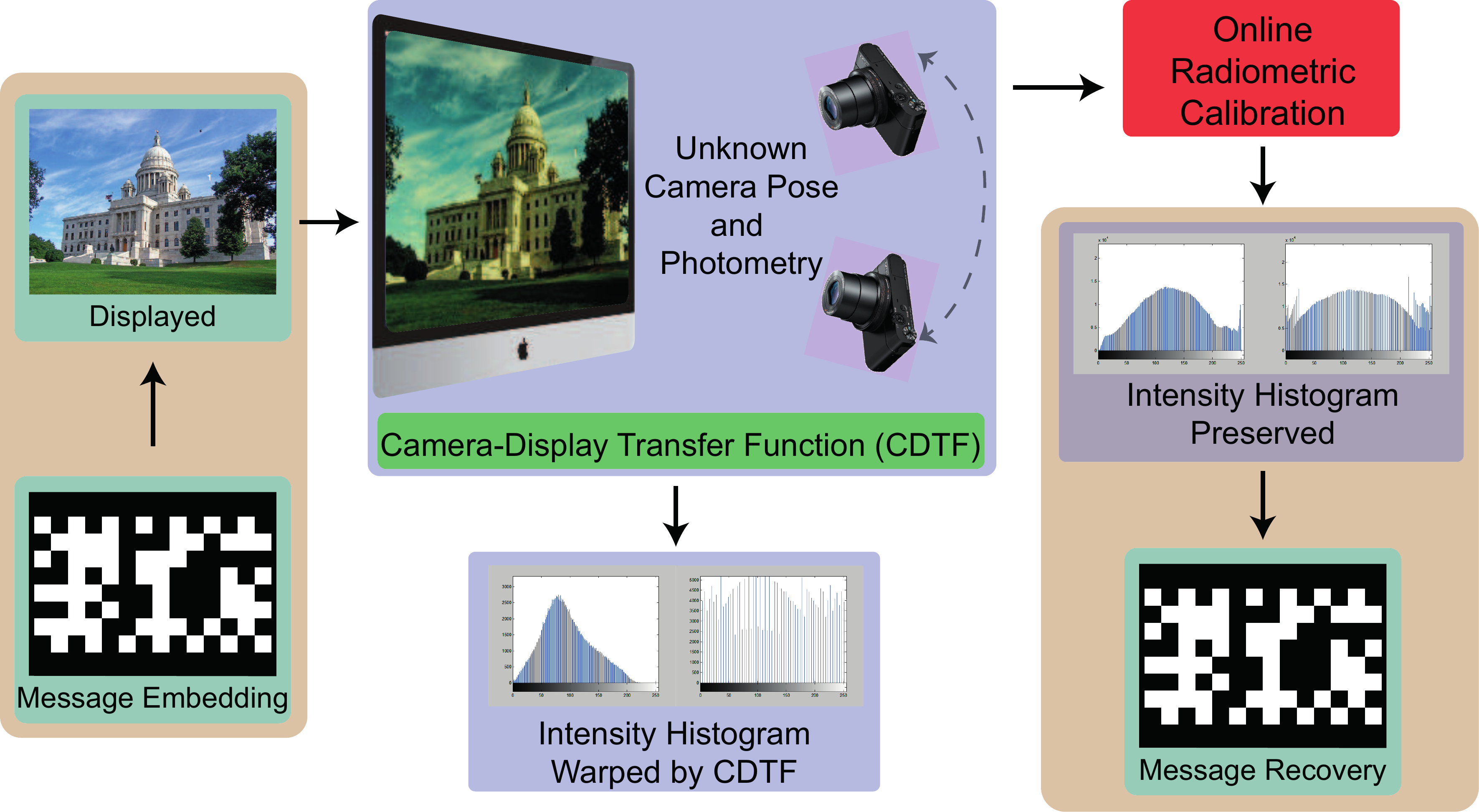}
	\end{center}
		\caption{The above flowchart illustrates the process by which Online Radiometric Calibration is used to estimate and negate the light-altering effects of the Camera-Display Transfer Function (CDTF) in camera-display communication. Variables such as camera pose, photometry, and hardware all have a significant effect on light signals passing from electronic display to camera. In each pair of intensity histograms shown above, the left represents an image's histogram before passing through the CDTF, and the right represents the histogram after the CDTF. Online Radiometric Calibration mitigates the distorting effects of the CDTF to preserve the image's histogram, enabling more accurate image recovery. } 	
\label{fig:ratex}
\end{figure}
%

%
\begin{comment}
\begin{figure}[t]
	\begin{center}
		\includegraphics[width=3in]{qrcodesmall.png}
	\end{center}
	\caption{QR code on a Times Square electronic billboard. The high contrast black-white pattern is relatively easy to detect, track and decode.  However, consider the more general task of encoding a message within an unknown and arbitrary image on an electronic display. The detection, tracking and decoding problems become significantly more challenging and interesting. } 		\label{fig:nycqrcode}
\end{figure}
\end{comment}
%

While traditional computer vision concentrates on objects that reflect environment lighting (passive scenes),  objects which emit  light, such as electronic displays, are increasingly common in modern scenes.  Unlike passive scenes,  {\it active scenes} can have intentional information that  must be detected and recovered.    For example, displays with QR codes \cite{ISO00} can be found in numerous locations such as shop windows and billboards.  However, QR-codes are very simple examples  because the bold, static pattern makes detection somewhat trivial.  The problem is more challenging from a computer vision point of view when the codes are not visible markers, but rather are hidden within a displayed image.  The displayed image is a light field, and decoding the message is an interesting problem in photometric modeling and computational photography.  
The paradigm has numerous applications because the electronic display and the camera can act as a communication channel where the display pixels are transmitters and the camera pixels are receivers \cite{Ashok10a}\cite{Ashok11}\cite{Varga11}.
Unlike hidden messaging in the digital domain,  prior work in real-world camera-display messaging is very limited.    
In this paper, we develop an optimal method for sending and retrieving hidden time-varying messages using  electronic displays and  cameras which accounts for the the characteristics of light emittance from the display.  We assume the electronic display has two simultaneous purposes: 1) the original display function such as advertising, maps, slides, or artwork;  2) the transmission of hidden time-varying messages.

\begin{figure}[t]
\centering
\includegraphics[width=3in]{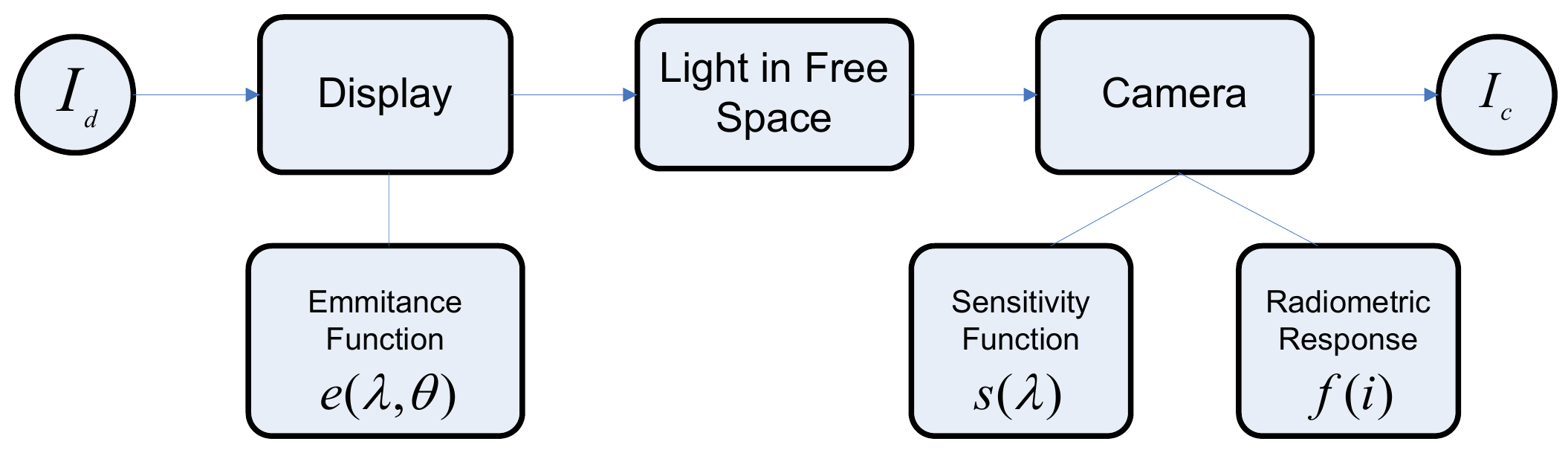}
\caption{Image Formation Pipeline: The image $\bf I_d$ is displayed by an electronic display with an emittance function $e$. The display is observed by a camera with sensitivity $s$ and radiometric response function $f$.  }
\label{fig:flowdiagram}
\end{figure}

When  light is emitted from a display, the resultant 3D light field has an intensity that depends on the angle of observation as well as the pixel value controlled by the display.   The emittance function of the electronic display is analogous to the BRDF (bidirectional reflectance distribution function) of a surface. This function characterizes the light radiating from a display pixel. It has a particular spectral shape that does not match the spectral sensitivity curve of the camera.   The effects of the display emittance function, the spectral sensitivity of the camera and the effect of camera viewing angle are all components of our photometric model for image formation as shown in Figure~\ref{fig:flowdiagram}.  Our approach does not require measurement or knowledge of the exact display emittance function.   Instead, we measure the entire system transfer function, as a {\it camera-display transfer function} (CDTF), which determines the captured pixel value as a function of the displayed pixel value.  By using frame-to-frame characterization of the CDTF, the method is independent of the particular choice of display and camera.

Interestingly,  while our overall goal has very strong similarities to the field of watermarking and steganography,  we present results that are novel and are aligned with the goals of computational photography.   Although watermarking literature has many hidden messaging methods, this area largely ignores the physics of illumination.
Display-camera messaging is fundamentally different from watermarking because each pixel of the image is a light source 
 that propagates in free space. Therefore, representations and methods that act only in the digital domain are not sufficient. 
 %Section~\ref{sec: related} discusses this issue in more detail. 

 The problem of understanding the relationship between the displayed pixel and the captured pixel is closely related
 to the area of radiometric calibration \cite{Mitsunaga99}\cite{Debevec97}\cite{Nayar00}.  In these methods,   a brightness transfer function characterizes the relationship between scene radiance and image pixel values.  The characterization of this function is done by measuring a range of scene radiances and the corresponding capture images pixels.   Our problem in camera-display messaging is similar but has important key differences. 
The CDTF is more complex than standard radiometric calibration because the system consists of both a display and a camera, each device adding its own nonlinearities. 
We can exploit the control of pixel intensities on the display and easily 
 capture the full range of input intensities. 
 However, the display emittance function  is typically dependent on the display viewing angle. Therefore, the  CDTF is dependent on camera pose. In a moving camera system, the CDTF must be estimated per frame; that is,  an online CDTF estimation is needed.
Furthermore,  this function varies spatially over the electronic display surface.  
%Examples of measured CDTFs that vary with viewing angle are shown in Figure~\ref{fig:cdtf_viewing}. Examples of the spatial variation of CDTF over the display surface are shown Figure~\ref{fig:cdtf_spatial}.

% Method comparison
\begin{figure*}[t]
  \centering
% \subfigure[Difference image]{\includegraphics[width=1.8in]{imgs/img_method_comp_diff.png}}
 % \subfigure[Thresholding]{\includegraphics[width=1.8in]{imgs/img_method_comp_thesholding.png}}
  % \subfigure[Our method]{\includegraphics[width=1.8in]{imgs/img_method_comp_our_method.png}}
  % needed an anti-aliasing filter ...
   \subfigure[Difference image]{\includegraphics[width=1.8in]{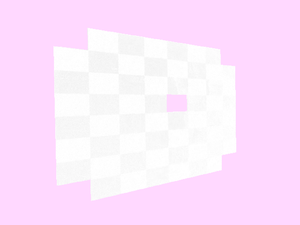}}
  \subfigure[Thresholding]{\includegraphics[width=1.8in]{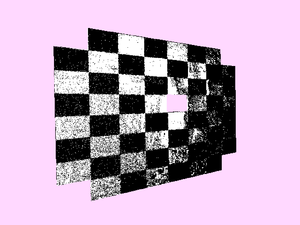}}
   \subfigure[Our method]{\includegraphics[width=1.8in]{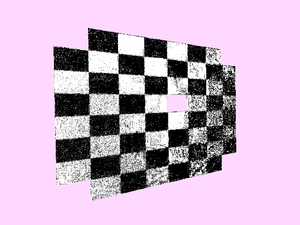}}
  \caption{Comparison of message recovery with a naive method and the proposed optimal method  (a) Difference of two consecutive frames in the captured sequence to reveal the transmitted message. (b) Naive method: Threshold the difference image by a constant (threshold $T=5$ for this example). (c) Optimal Method: Bits are classified by a simultaneous radiometric calibration and support vector machine classifier. }
  \label{fig:thresholdoptimal}       % Give a unique label
\end{figure*}

% Ratex patch insertion to ensure inconspicuous presence can be done by a simple histogram matching algorithm which transforms the brightness values of certain image regions to match a desired histogram. 

We show that the two-part problem of online radiometric calibration and accurate message retrieval can be structured as an optimization problem.  This leads to  the primary contribution of the paper. We present an elegant problem formulation where the photometric modeling leads to {\it physically-motivated kernel functions} that are used with a support vector machine classifier.   We show that calibration and message bit classification can be done simultaneously and the resulting optimization algorithm
operates in four dimensional space and is convex.   The algorithm is a novel method for {\it online optimal radiometric calibration}  (OORC) that enables accurate camera-display messaging.  
An example message recovery result is shown in Figure~\ref{fig:thresholdoptimal}. 
Our experimental results show that accuracy levels for message recovery can improve from as low as 40-60\% to higher than 90\%  using our approach when compared to either no calibration, or sequential calibration followed message recovery.  
%The accuracy of information retrieval  can support interesting applications where displayed images can act as communications channels \cite{Ashok11}\cite{Ashok11a}\cite{Varga11}.
For evaluation of results, 9 different combinations of displays and cameras are used with 15 different image sequences, for multiple embedded intensity values, and multiple camera-display view angles.

The standard problem of radiometric calibration is solved by varying exposure so that a range of scene radiance can be measured. For CDTF estimation, textured patches are placed within the display image that have intensity variation over the full range of display brightness values.  These patches  can be placed in inconspicuous regions of the display image or in corners.  We use the term {\it ratex patch} to refer to these radiometric calibration texture patches.  The ratex  patches are not used as part of the hidden message. Multiple ratex patches can be used to find a spatially varying CDTF.  The ratex patches have the advantage that they  are perceptually acceptable, they represent the entire range of  gray-scale intensity variation, and they can be distributed spatially.  Furthermore, these patches are used for support vector machine training as described in Section~\ref{sec:methods}.

Additionally, we introduce a method of radiometric calibration that employs visually non-disruptive ``hidden ratex'' mapping. Rather than directly measuring the effect that the CDTF has on known intensity values, we are able to model the CDTF based on changes to a known frequency distribution of intensity values. Radiometric calibration with hidden ratex produces a distribution-driven intensity mapping that mitigates the photometric effects of the CDTF for simple message recovery.

 The contributions of the paper can be summarized as follows:
 1) A new optimal online radiometric calibration with simultaneous message recovery, cast as a convex optimization problem; 2) photometric model of the camera display transfer function; 3) the use of ratex patches to provide continual calibration information as a practical method for online calibration; 4) the use of distribution-driven intensity mapping as a practical method for visually non-disruptive online calibration.

\section{Related Work}\label{sec:related}

\paragraph{Watermarking}
In developing a system where cameras and displays can communicate under real world conditions, the initial expectation was that existing watermarking  techniques could be used directly.  Certainly the work in this field is extensive and has a long history with numerous surveys compiled \cite{Cheddad10} \cite{Wayner09} \cite{Potdar05} \cite{Cox02} \cite{Johnson01} \cite{Petitcolas99}.
%Petitcolas99b
Surprisingly, existing methods are not directly applicable to our problem.  In the field of  watermarking,  a fixed image or mark is embedded in an image often with the goal of identifying fraudulent copies of a video, image or document.   Existing work emphasizes almost exclusively the digital domain and  does not account for the effect of illumination in the image formation process in real world scenes. 
In the digital domain, neglecting the physics of illumination is quite reasonable; however, for camera-display messaging, illumination plays a central role.

From a computer vision point of view, the imaging process can be divided into two main components: photometry and geometry.  The geometric aspects of image formation have been addressed to some extent in the watermarking community, and many techniques have been developed for robustness to geometric changes during the imaging process such as scaling, rotations, translations and general homography transformations \cite{Dong05} \cite{Sangeetha09} \cite{Dugelay06} \cite{Wang08} \cite{Lin01} \cite{Potdar05} \cite{Seo06}. 
However, the {\it photometry} of imaging has largely been ignored. 
The rare mention of photometric effects \cite{Zou09} \cite{Yang06} in the watermarking literature doesn't define photometry with respect to illumination; instead photometric effects are defined as ``lossy compression,
denoising, noise addition and lowpass filtering''.
 In fact,  photometric attacks are sometimes defined as jpeg compression \cite{Dugelay06}.

 \paragraph{ Radiometric Calibration} 
 Ideally, we consider the pixel-values in a camera image to be a measurement of light incident on the image plane sensor.  It is well known that the relationship is typically nonlinear.  
Radiometric calibration methods have been developed to estimate the camera response function that converts irradiance to pixel values. In measuring a camera response,  a series of known brightness values are measured along with the corresponding pixel values. In general, having such ground truth brightness is quite difficult. 
 The classic method \cite{Debevec97} uses  multiple exposure values instead. The light intensity on the sensor is a linear function of the time of exposure, so known exposure times enables  ground truth light intensity.  This exposure-based method is used in several  radiometric calibration methods 
 \cite{Mitsunaga99} \cite{Nayar00} \cite{Debevec97} \cite{Mann95} \cite{Kim08}.  
 Our goal for the display-camera system is related to radiometric calibration, yet different in significant ways.  We are interested not just in a system that converts scene radiance to pixels (the camera), but also converts from pixel to scene radiance (the display) so that the whole camera-display system is a function that maps a color value at the display to a color value at the camera.  
  
The camera response in radiometric calibration is either estimated as a full mapping where $i_{out}$ is specified for every $i_{in}$ or as an analytic function $g(i_{in})$. 
 Several authors \cite{Mitsunaga99} \cite{Chakrabarti09} \cite{Lee13} use polynomials  to model the radiometric response function. Similarly, we have found that fourth order polynomials 
 %with a monotonic brightness constraint 
 can be used for modeling the inverse display-camera transfer function. 
 The dependence on color is  typically modeled by considering each channel independently \cite{Mitsunaga99} \cite{Nayar00} \cite{Debevec97} \cite{Grossberg02} .  Interestingly, 
although more complex color models have been developed \cite{Kim12} \cite{Lin11} \cite{Xiong12}, we have found the independent channel approach suitable for the display-camera representation where the optimality criterion is accurate message recovery. 

Existing radiometric calibration methods are developed for cameras, not camera-display systems. Therefore,  display emittance function is not part of the system to be calibrated.  However, for the camera-display transfer function, this component plays an important role. 
We do not use the measured display emittance function explicitly, but since the CDTF is view dependent and the camera can move,
our approach is to perform radiometric calibration per frame, by the insertion of radiometric calibration patches (ratex patches).

\paragraph{Other Methods for Camera-Display Communication} %Note that prior work in Camera display messaging does not include radiometric calibration\\
 Camera-display communications have precedent in the computer vision community, but existing methods differ from our proposed approach. For example,
researchers on the  Bokode project \cite{Mohan09} presented a system using an invisible message, however the message is a fixed symbol, not a time-varying message. Invisible QR codes were addressed in \cite{Kamijo08}, but these QR-codes are fixed.  Similarly, traditional watermark approaches typically contained fixed messages.
LCD-camera communications is presented in
 \cite{Perli10} with a time-varying message, but the camera is in a fixed position with respect to the display. Consequently,  the  electronic display is not detected, tracked or segmented from the background. Furthermore, the transmitted signal is not hidden in this work. 
Recent work has been done in high speed visible light communications \cite{Vucic10}, but this work does not utilize existing displays and cameras and requires specialized hardware and LED devices. 
Time-of-flight cameras have recently been used for phase-based communication \cite{yuan2014phase}, but these methods require special hardware.
Interest in camera-display messaging is also shared in the mobile communications domain.
COBRA, RDCode, and Strata have developed 2D barcode schemes designed to address the challenges of low-resolution and slow shutter speeds typically present in smartphone cameras \cite{hao2012cobra} \cite{wang2014enhancing} \cite{hu2014strata}.
Likewise, Lightsync has targeted synchronization challenges with low frequency cameras. \cite{hu2013lightsync}.

%  
%  % Wenjia insert references for the tracking methods you are using.
%  %
\begin{comment}
\paragraph{Steganography}
Our approach can be called {\it photographic steganography } to emphasize that the approach is a blend of computational photography and steganography.   In this work, the meaning of steganography follows the definition ``message hiding'' instead of ``secret message''.  The message that is embedded in the image is not 
meant to be secret in the strict sense, instead the message is to be hidden so that the image on the electronic display is perceptually unchanged by the message. 
Traditionally, the field of steganography in digital multimedia emphasizes secret messages  \cite{Avcibas03}\cite{Petitcolas99}\cite{Cheddad10}, but
our use of the word is an adaptation for the field of computer vision  where the camera is intended to recover the message that is hidden to the eye. 
\end{comment}

%%%%%%%%%% SYSTEM PROPERTIES %%%%%%%%%%%
%----------------------------------------------------------------------------
%----------------------------------------------------------------------------
\section{System Properties}\label{sec:sys_properties}

% definition
In our proposed camera-display communication system, pixel values from the display are inputs, while captured intensities from the camera are output. We denote the mapping from displayed intensities to captured ones as {\it Camera-Display Transfer Function} (CDTF).
% method 
In this section, we motivate the need for online radiometric calibration by briefly analyzing factors that commonly influence the CDTF.

\subsection{Display Emittance Variation}
\begin{figure}[h]
\centering
% row 1
\subfigure{\label{fig:display_Samsung_bmp}\includegraphics[width=0.14\textwidth]{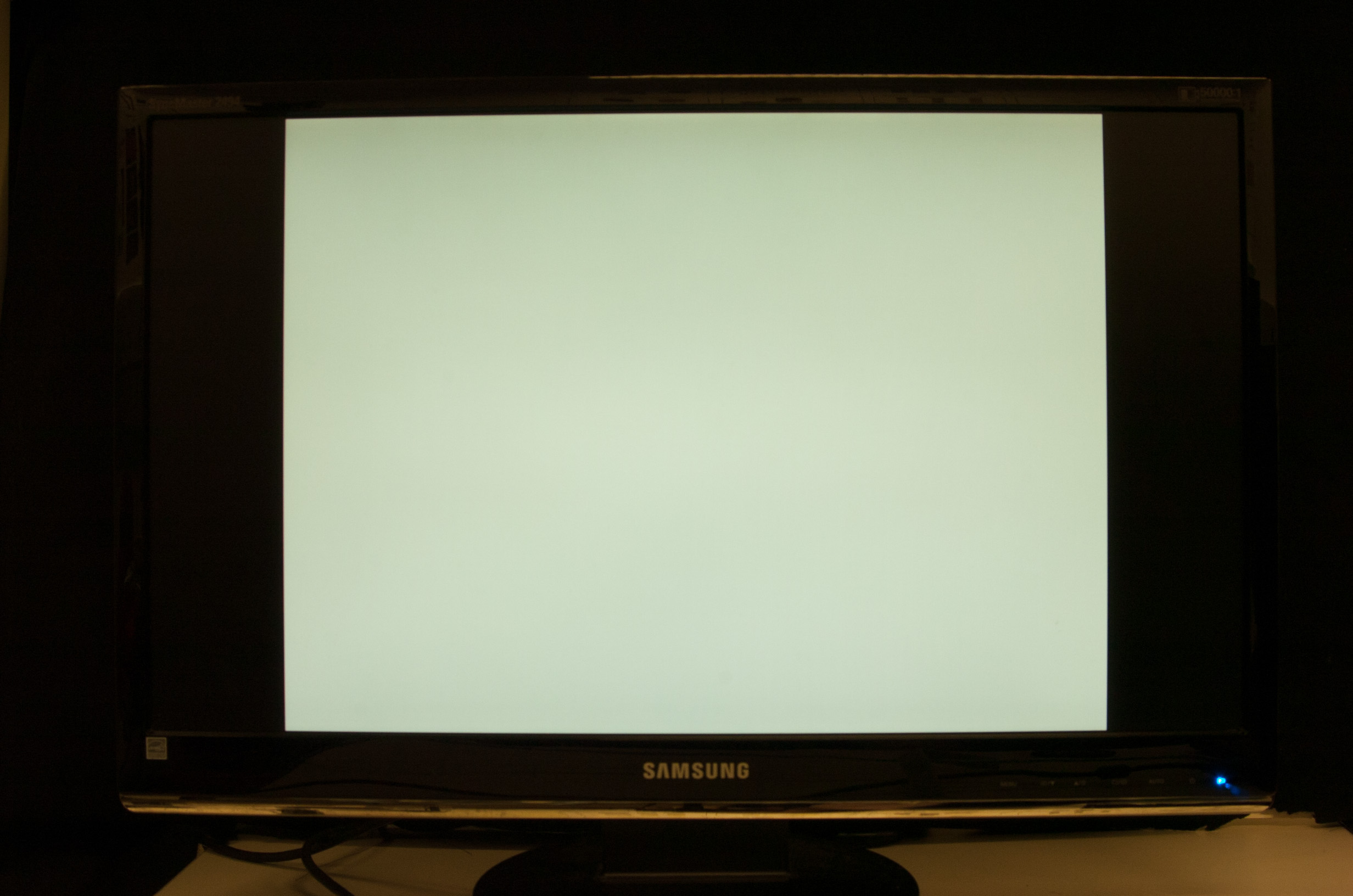}} 
\subfigure{\label{fig:display_LG_bmp}\includegraphics[width=0.14\textwidth]{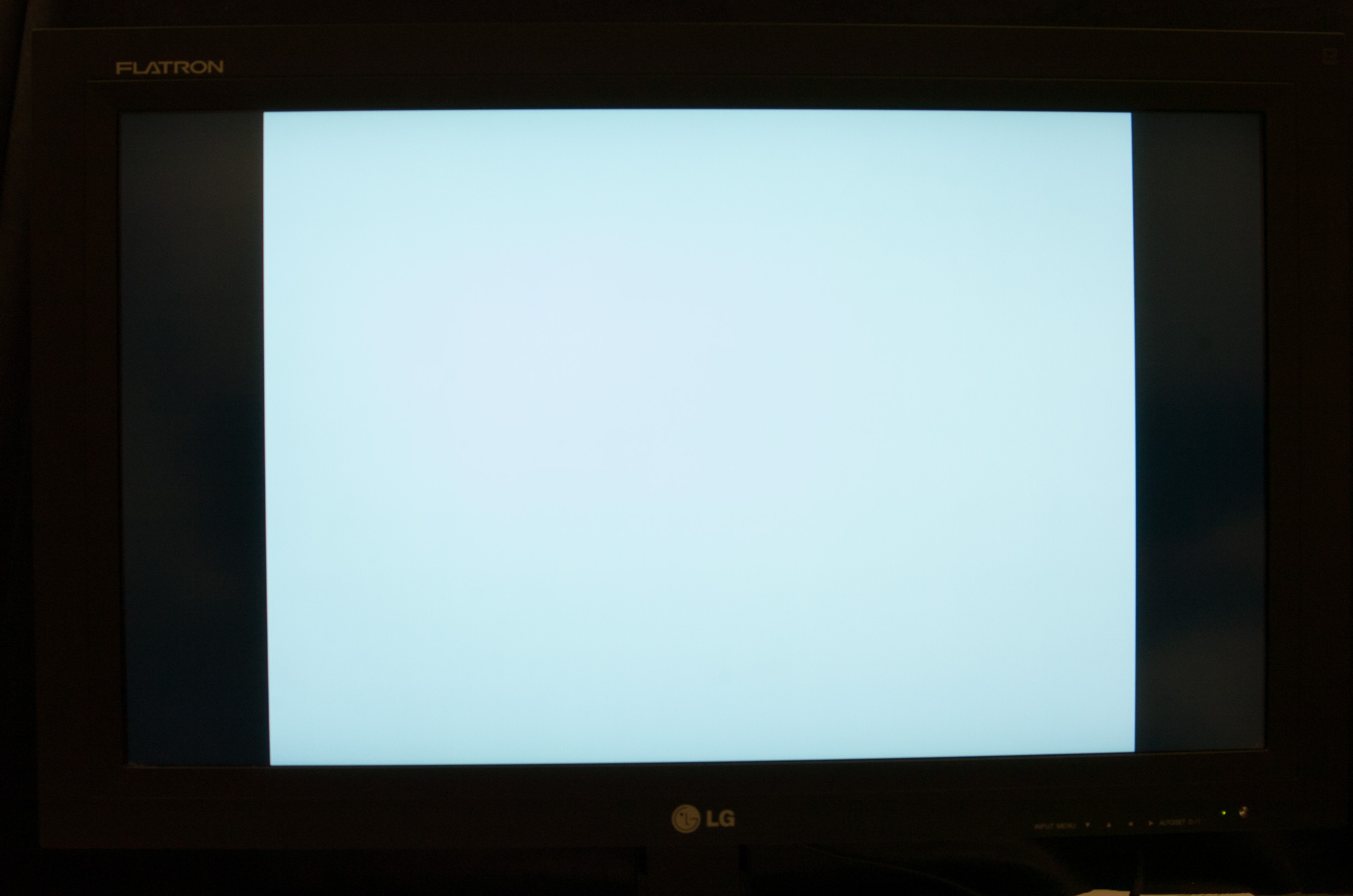}} 
\subfigure{\label{fig:display_iMac_bmp}\includegraphics[width=0.14\textwidth]{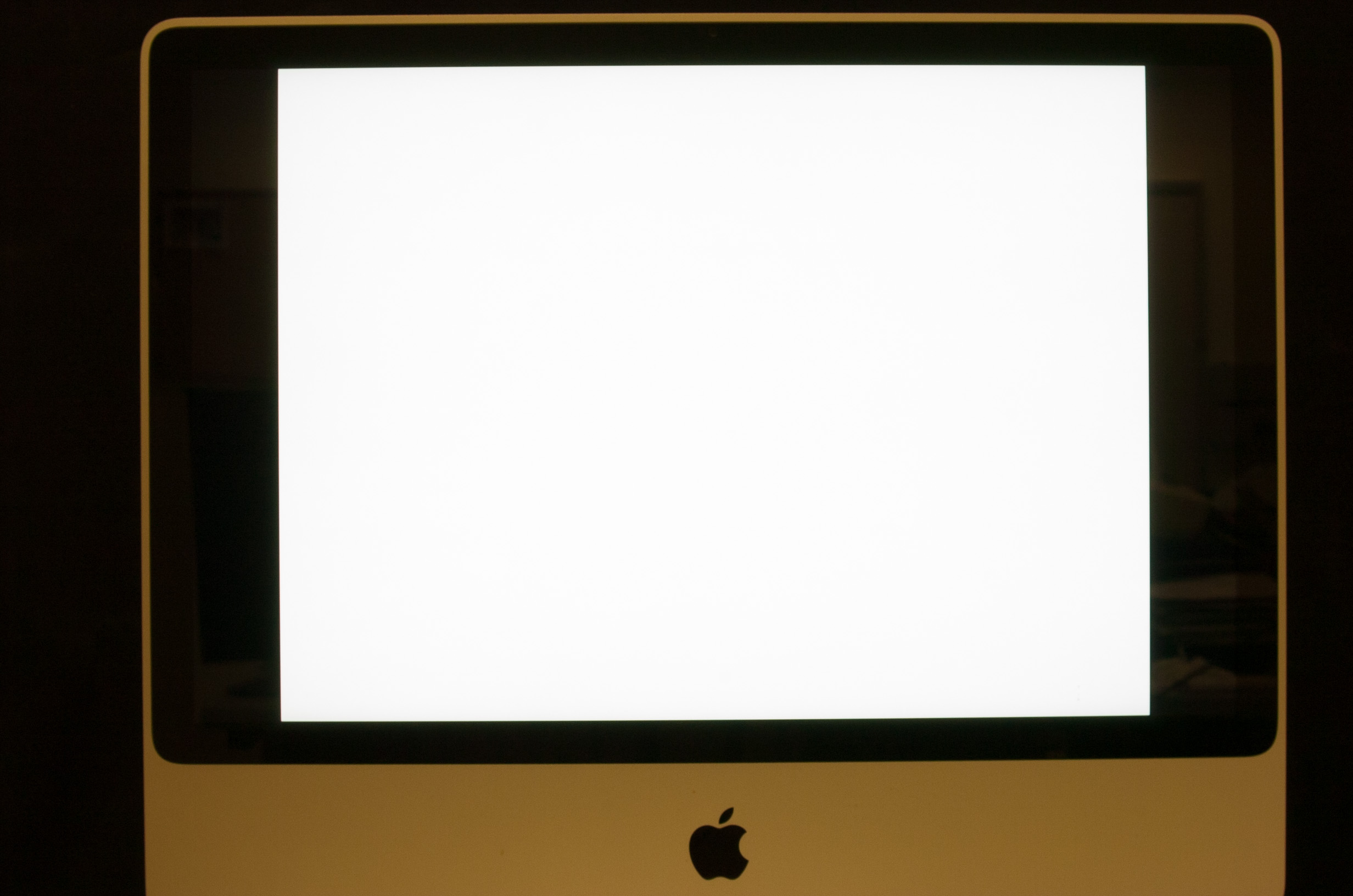}}
% row 2
\subfigure[Samsung]{\label{fig:display_Samsung_eps}\includegraphics[width=0.15\textwidth]{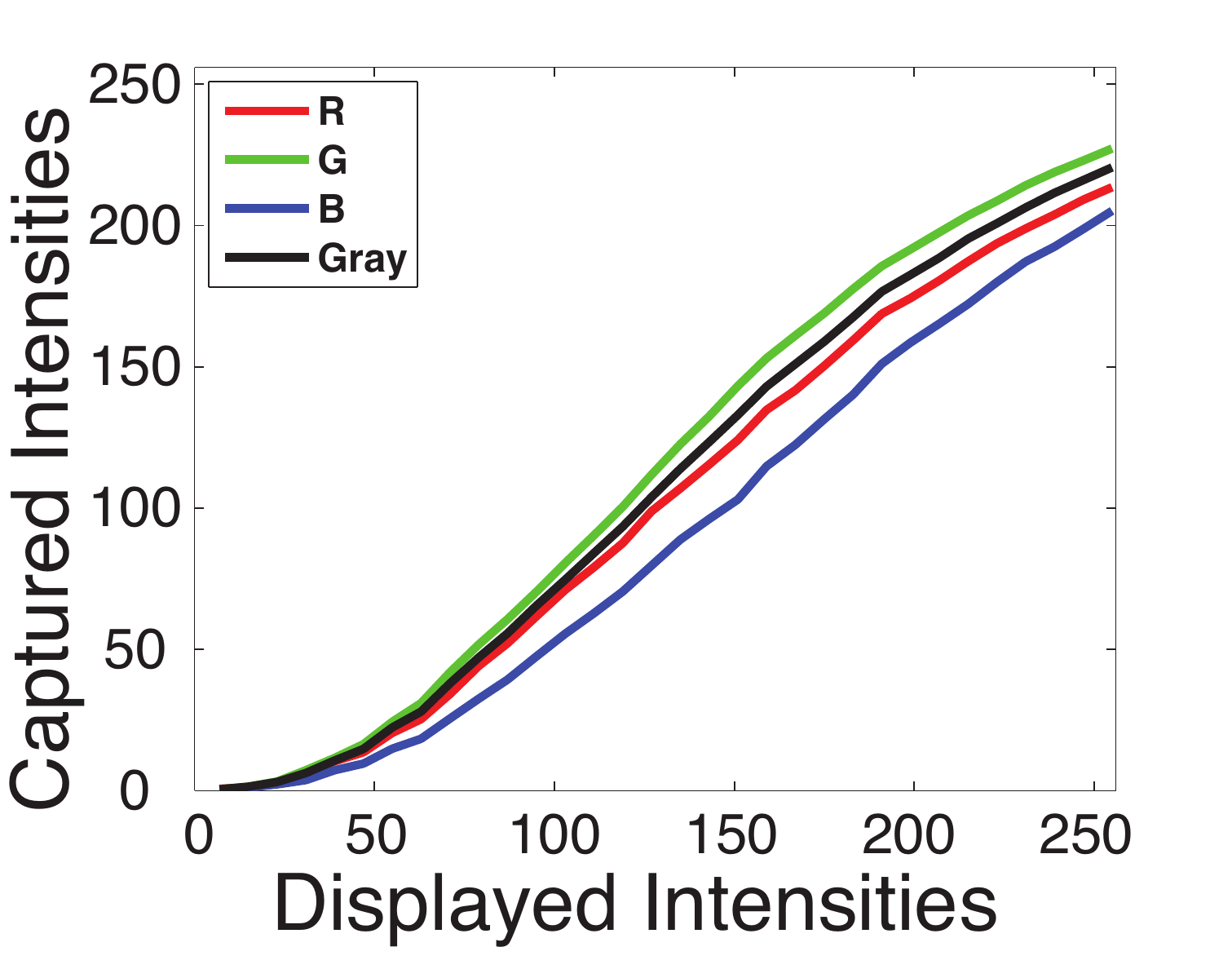}\setcounter{subfigure}{1}} 
\subfigure[LG]{\label{fig:display_LG_eps}\includegraphics[width=0.15\textwidth]{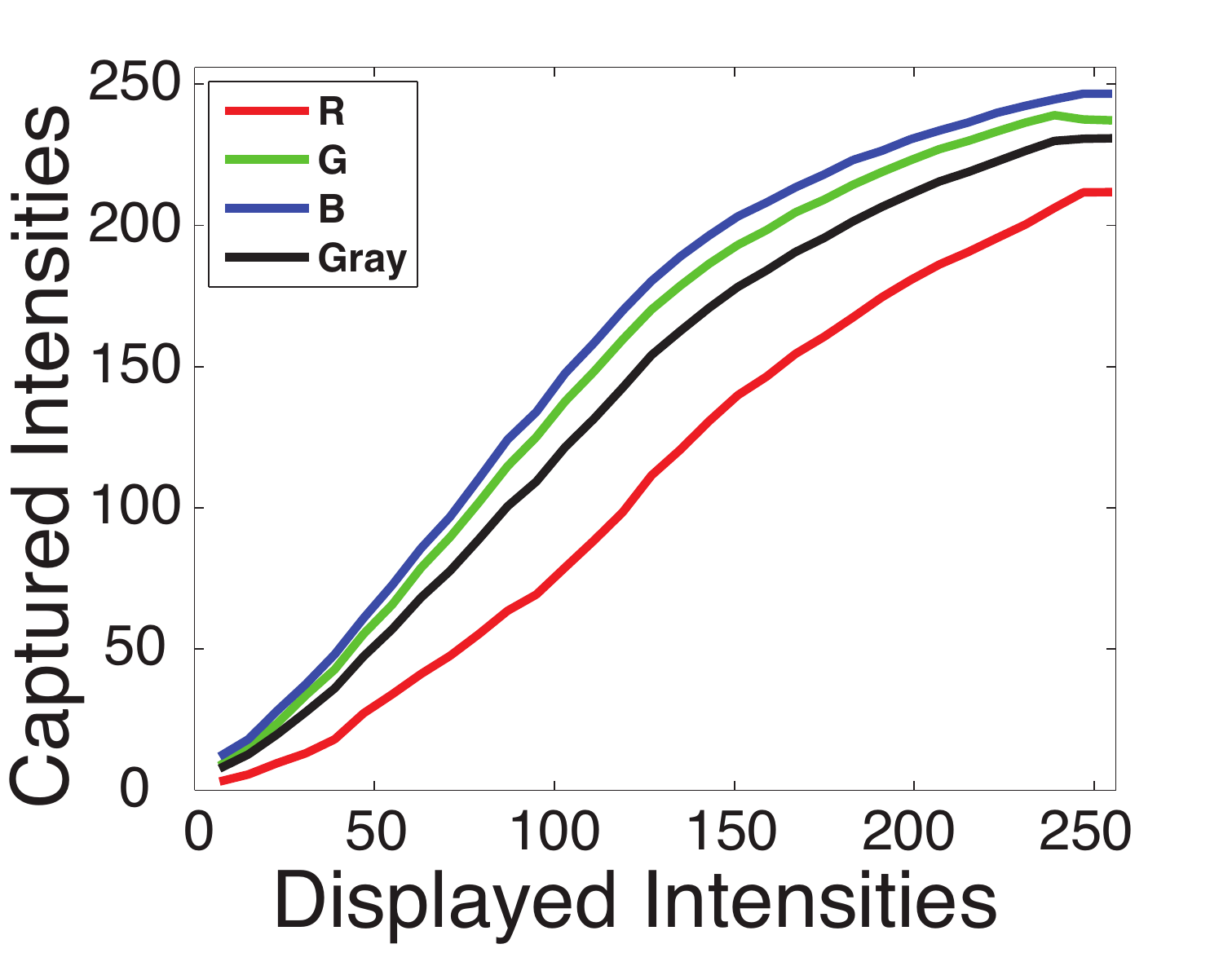}} 
\subfigure[iMac]{\label{fig:display_iMac_eps}\includegraphics[width=0.15\textwidth]{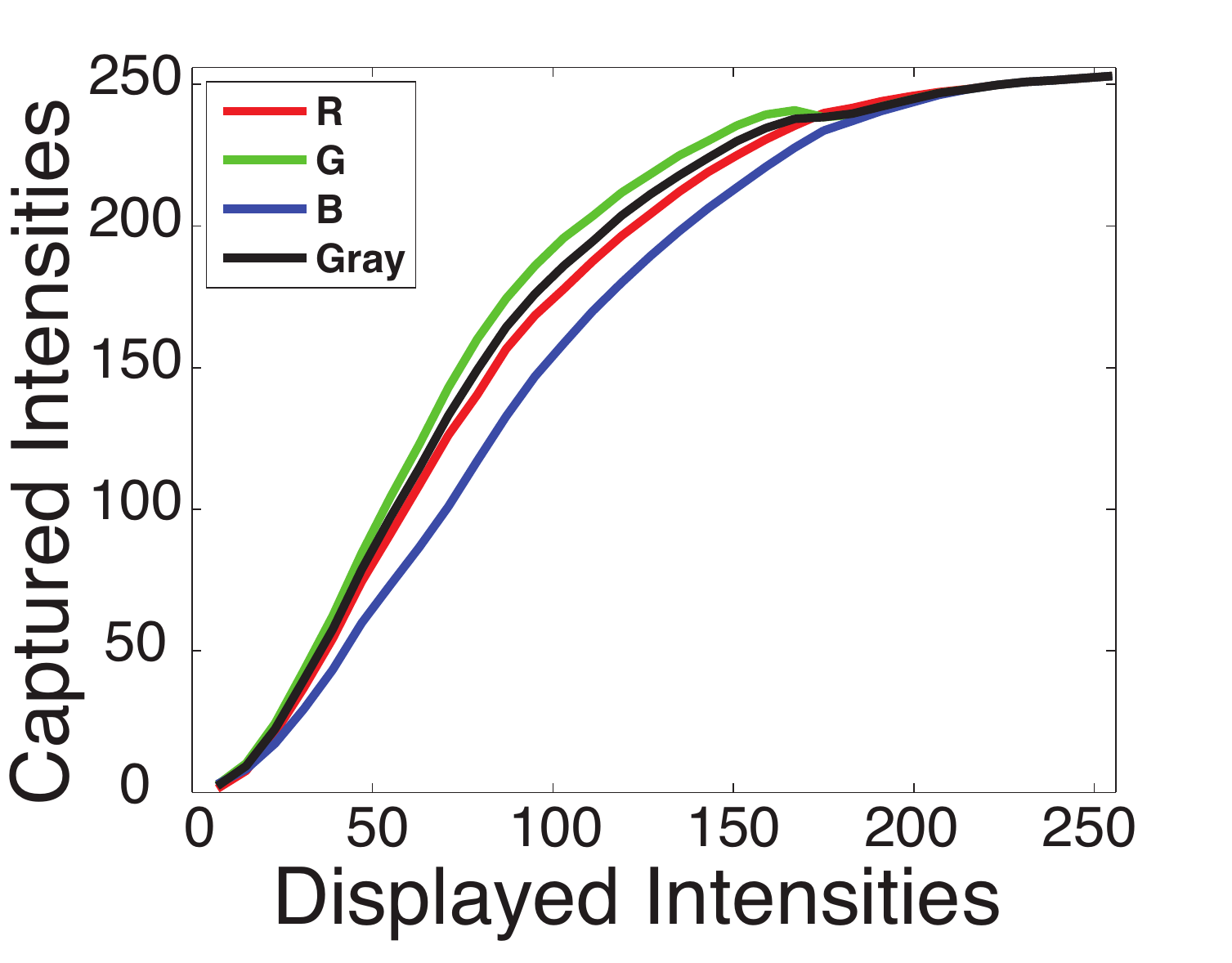}}
\caption{{\bf Variance of Light Output among Displays}. An SLR camera captured a range of grayscale [0,255] intensity values produced by 3 different LCDs. These 3 CDTF curves highlight the dramatic difference in the light emmitance function for different displays, particularly the LG.}
\label{fig:camera_3displays}      
\end{figure}

Displays vary widely in brightness, hue, white balance, contrast and many other parameters that will influence the appearance of light. To affirm this hypothesis, an SLR camera with fixed parameters observes 3 displays and models the CDTF for each one. See Samsung in Fig.~\ref{fig:display_Samsung_bmp}, LG in Fig.~\ref{fig:display_LG_bmp}, and iMac~\ref{fig:display_iMac_bmp}. Although each display is tuned to the same parameters, including contrast and RGB values, each display produces a unique CDTF. 

% pic: the same camera for 3 displays

%----------------------------------------------------------------------------
% observation angles
\subsection{Observation Angles}
\begin{figure}[h]
\centering
% row 1
\subfigure{\label{fig:angle1_Nikon_Sam_bmp}\includegraphics[width=0.14\textwidth]{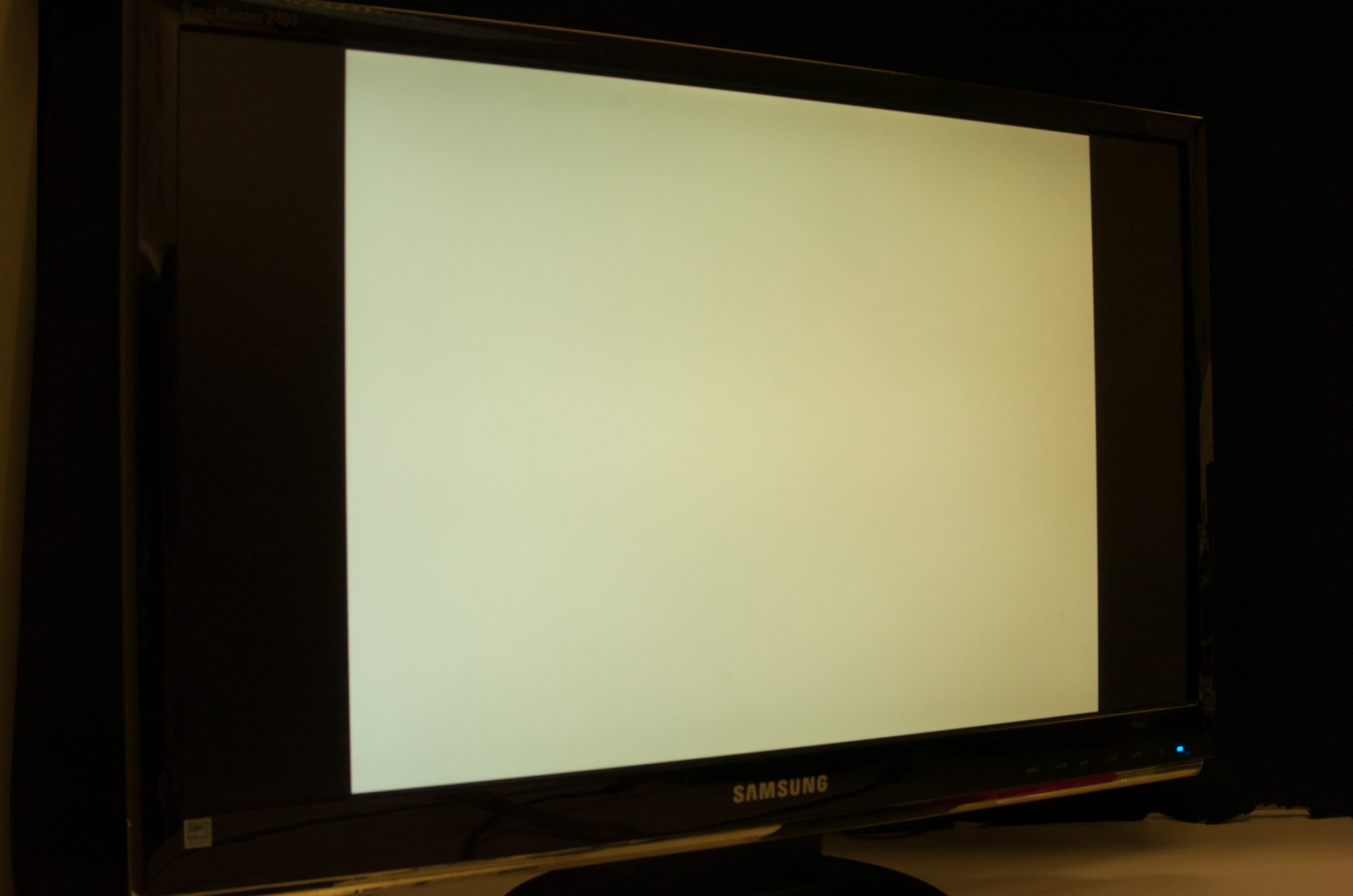}} 
\subfigure{\label{fig:angle2_Nikon_Sam_bm}\includegraphics[width=0.14\textwidth]{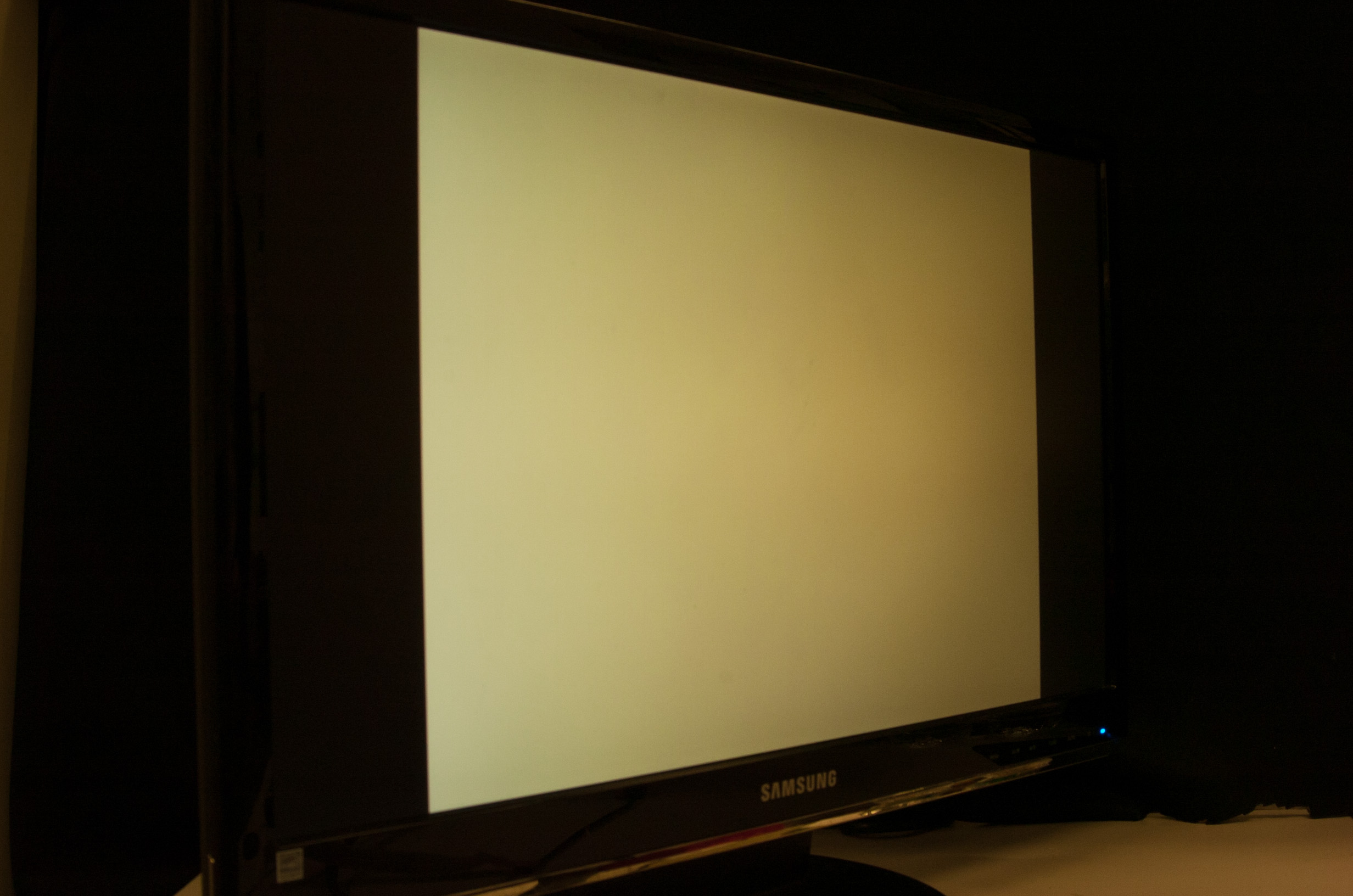}} 
\subfigure{\label{fig:angle3_Nikon_Sam_bm}\includegraphics[width=0.14\textwidth]{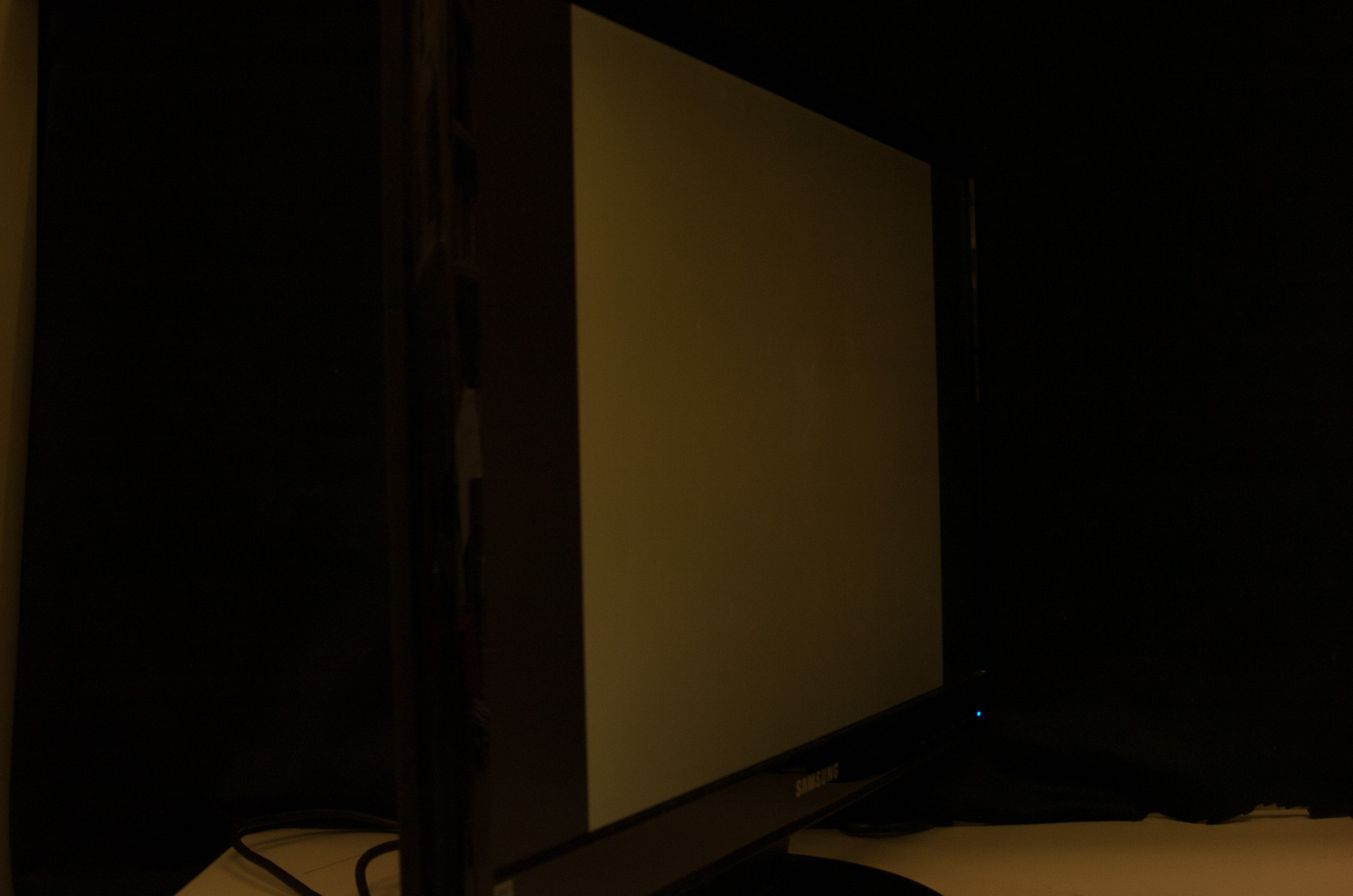}}
% row 2
\subfigure[30 \degree]{\label{fig:angle1_Nikon_Sam_eps}\includegraphics[width=0.15\textwidth]{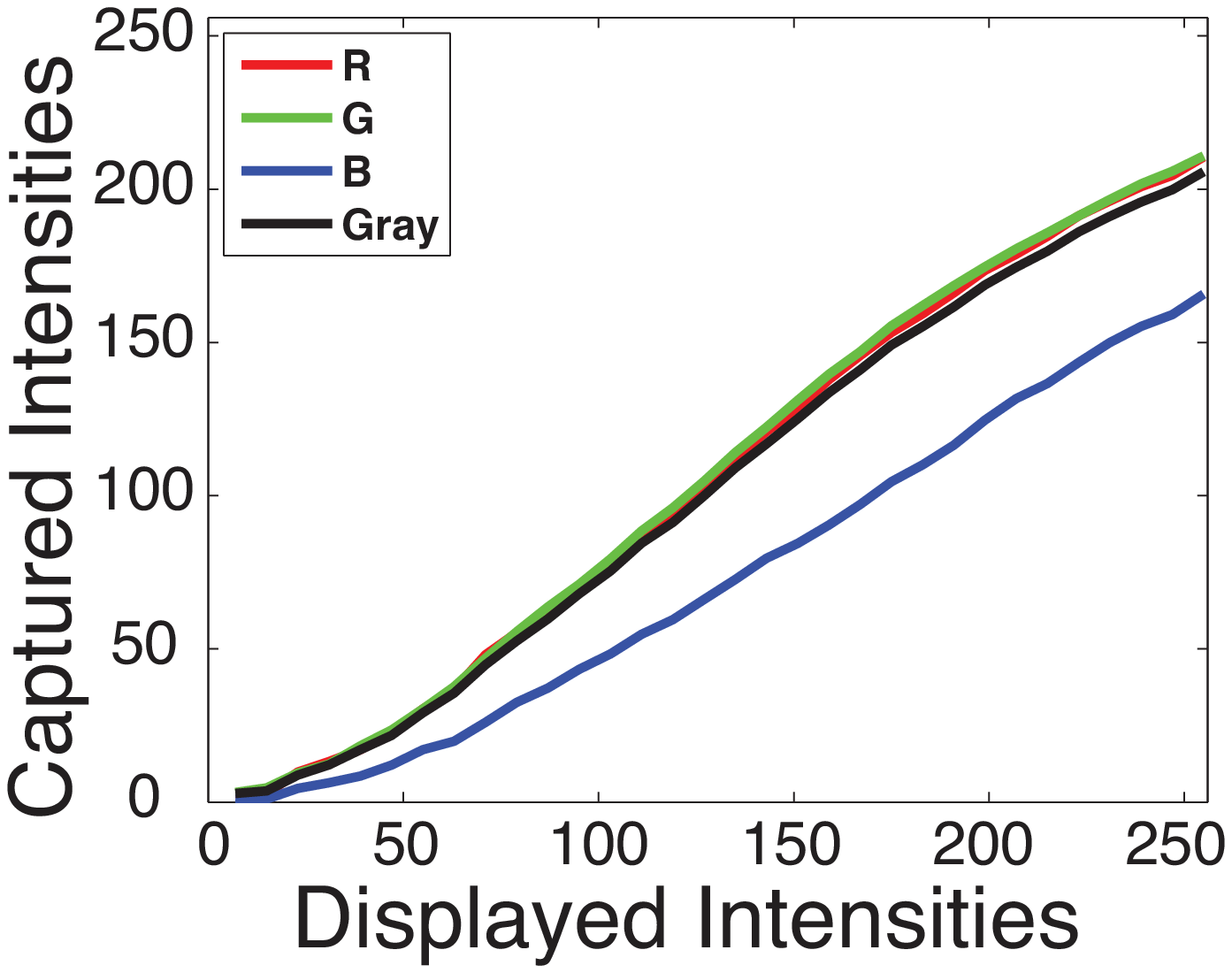}\setcounter{subfigure}{1}} 
\subfigure[45 \degree]{\label{fig:angle2_Nikon_Sam_eps}\includegraphics[width=0.15\textwidth]{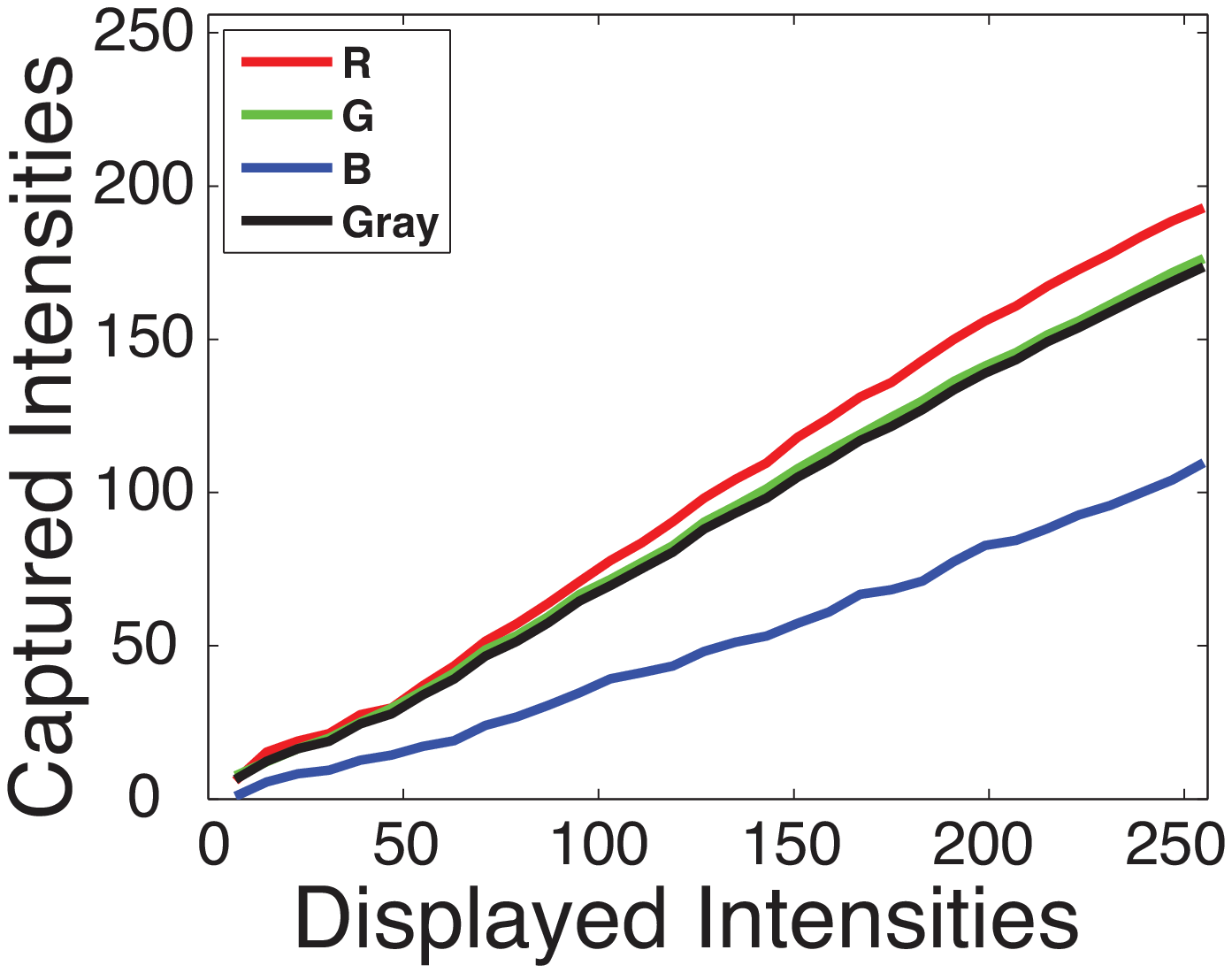}} 
\subfigure[60 \degree]{\label{fig:angle3_Nikon_Sam_eps}\includegraphics[width=0.15\textwidth]{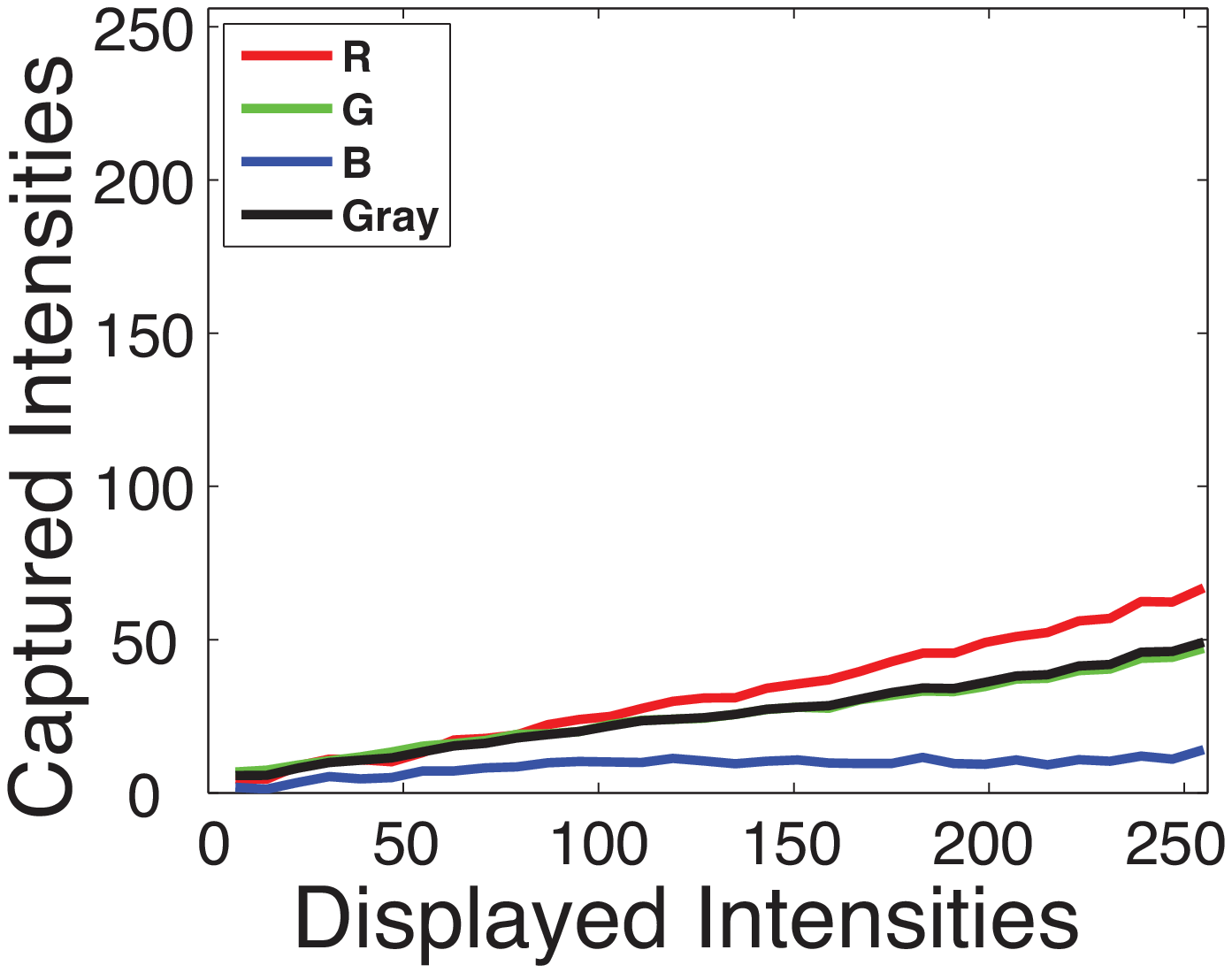}}
\caption{{\bf Influence of observation angles}. Using the Nikon-Samsung pair, a range of grayscale [0, 255] values were displayed and captured from a set of different observation angles. As observation angle became more oblique, the captured light intensity sharply decreased. Therefore, observation angle has a dramatic, nonlinear effect on CDTF. }
\label{fig:angles}      
\end{figure}

Displays do not emit light in all directions with the same power level. Therefore CDTF is also sensitive to observation angles. To verify this hypothesis, an experiment was performed where an SLR camera captured the light intensity produced by a computer display from multiple angles. The results in Fig.~\ref{fig:angles} show that more oblique observation angles yield lower captured pixels intensities. Moreover, there is a nonlinear relationship between captured light intensity and viewing angle. 

% pic: observation angles

%Histograms!
% pic: oblique view
\begin{figure}[h]
  \centering
% row 1
\subfigure{\label{fig:pos1_Nikon_Sam_bmp_o}\includegraphics[width=0.14\textwidth]{Angle1_Nikon_Sam.jpg}} 
\subfigure{\label{fig:pos2_Nikon_Sam_bmp_o}\includegraphics[width=0.14\textwidth]{Angle2_Nikon_Sam.jpg}} 
\subfigure{\label{fig:pos3_Nikon_Sam_bmp_o}\includegraphics[width=0.14\textwidth]{Angle3_Nikon_Sam.jpg}}
% row 2
\subfigure[30 \degree]{\label{fig:pos1_Nikon_Sam_hist_o_eps}\includegraphics[width=0.15\textwidth]{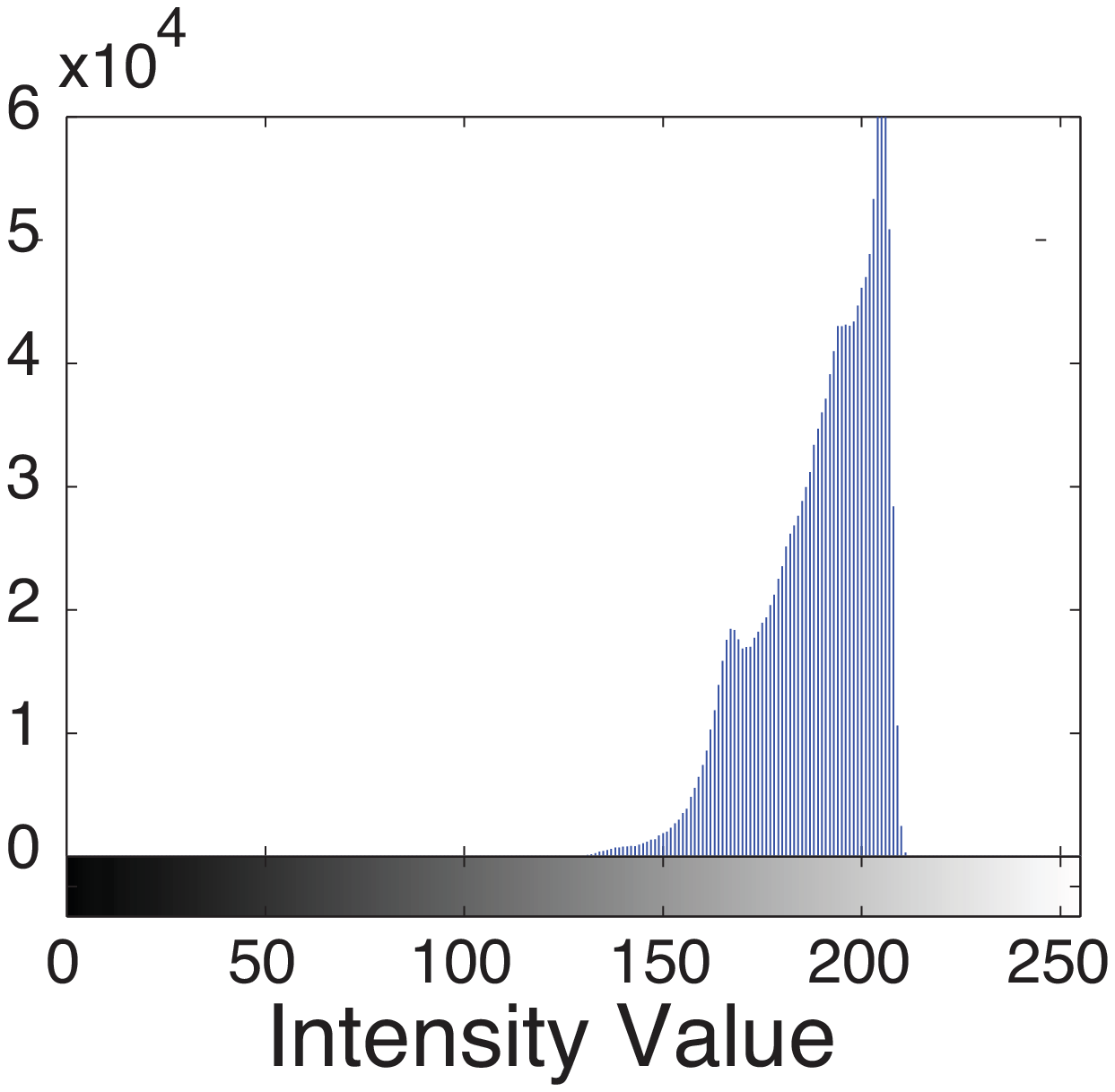}\setcounter{subfigure}{1}} 
\subfigure[45 \degree]{\label{fig:pos2_Nikon_Sam_hist_o_eps}\includegraphics[width=0.15\textwidth]{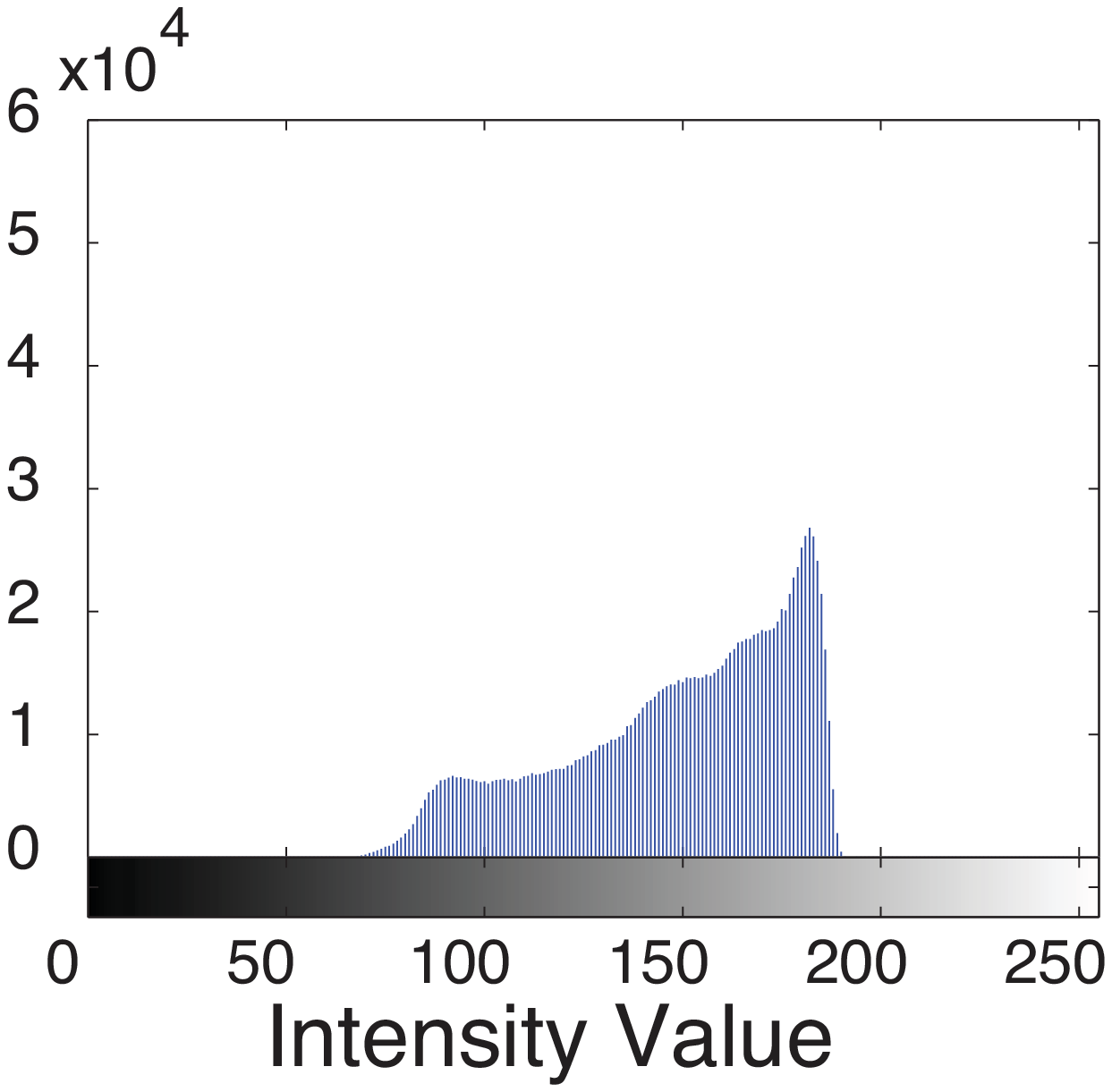}} 
\subfigure[60 \degree]{\label{fig:pos3_Nikon_Sam_hist_o_eps}\includegraphics[width=0.15\textwidth]{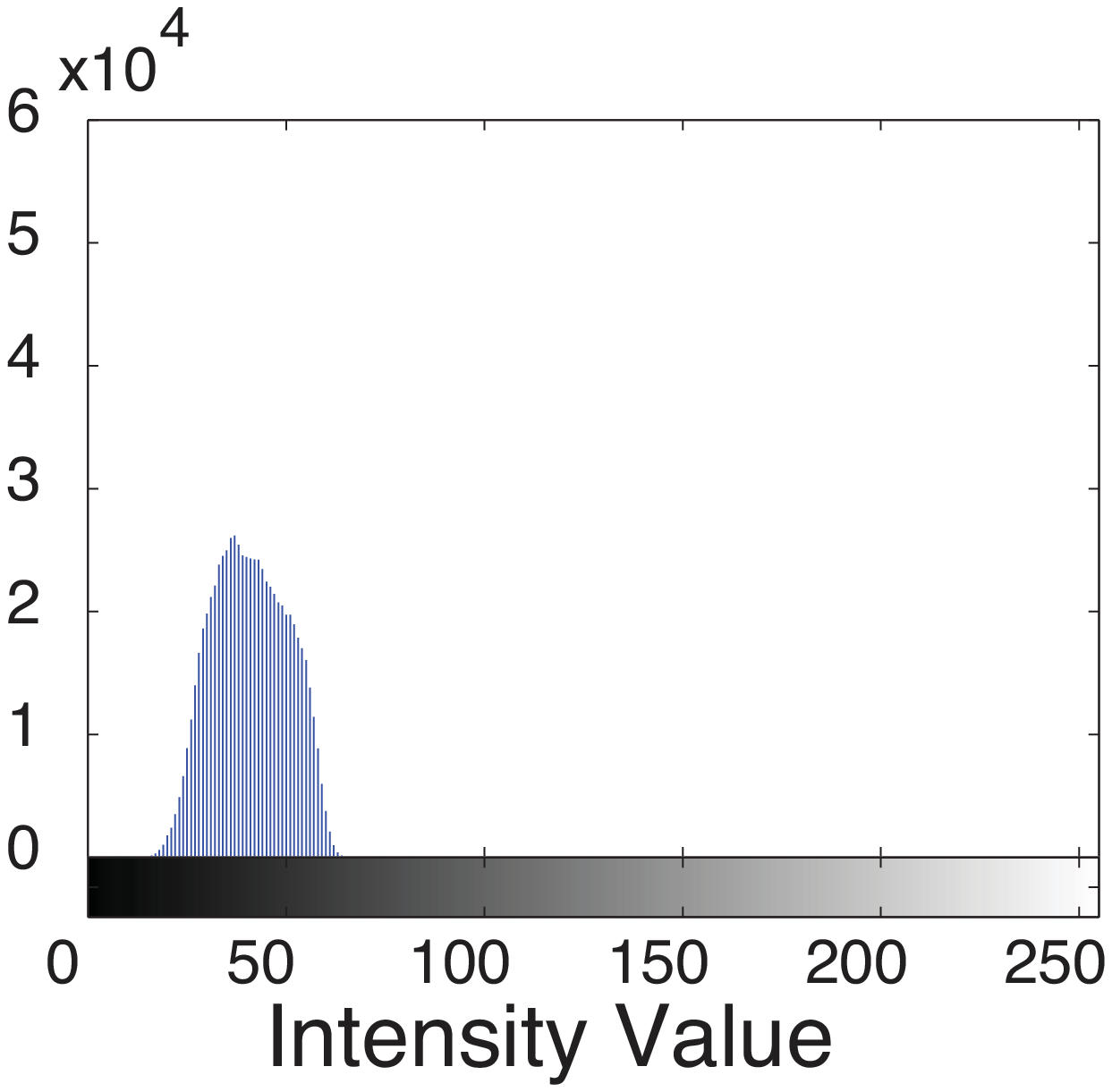}}
  \caption{{\bf Histograms of intensities across the display.} Notice as observation angle changes, so does the frequency distribution of captured intensities. If the intensity distribution (histogram) of the displayed image was known, an observer can estimate the CDTF.}
  \label{fig:spa_toning_o_hist}      
\end{figure}

\section{Methods}
\label{sec:methods}
\subsection{Photometry of Display-Camera systems}
\label{sec:photometric}

\begin{comment}
\begin{figure}[t]
	\begin{center}
	%	\subfigure[0$^{\circ}$]{	
			\includegraphics[width=0.075\textwidth]{angle90.png}
	%		\label{fig:0degree}}
	%	\subfigure[30$^{\circ}$]{
			\includegraphics[width=0.075\textwidth]{angle60.png}
	%		\label{fig:30degree}}
	%	\subfigure[50$^{\circ}$]{	
			\includegraphics[width=0.075\textwidth]{angle40.png}
	%		\label{fig:50degree}}
	%	\subfigure[60$^{\circ}$]{	
			\includegraphics[width=0.075\textwidth]{angle30.png}
	%		\label{fig:60degree}}
	%	\subfigure[65$^{\circ}$]{
			\includegraphics[width=0.075\textwidth]{angle25.png}
	%		\label{fig:65degree}}
	%	\subfigure[70$^{\circ}$]{	
			\includegraphics[width=0.075\textwidth]{angle20.png}
	%		\label{fig:70degree}}
	\end{center}
	\caption{Captured intensity from an LCD monitor decreases with viewing angle (from left to right), demonstrating view-dependent display emittance function.}
	\label{fig:LCDfalloff}
\end{figure}
\end{comment}

The captured image ${\bf I_c}$ from the camera viewing the electronic display image ${\bf I_d}$ can be modeled using the image formation pipeline in Figure~\ref{fig:flowdiagram}.
First, consider a particular pixel within the display image ${\bf I_d}$ with  red, blue and green components given by  ${ \bf \rho} = (\rho_r,\rho_g,\rho_b)$.
The captured image ${\bf I_c}$ at the camera has three color components $(I_c^r,I_c^g,I_c^b)$, however there is no one-to-one correspondence between the color channels of the camera sensitivity function and the electronic display emittance function.
When the monitor displays  the value $(\rho_r,\rho_g,\rho_b)$ at a pixel, it is emitting light in a manner governed by its emittance function and modulated by $\rho$. 
 The monitor emittance function ${\bf e}$  is typically a function of the viewing angle   ${\bf \theta}=(\theta_v,\phi_v)$ comprised of a polar and azimuthal component.
 For example, the emittance function of an LCD monitor has a large decrease in intensity with polar angle (see Figure~\ref{fig:spa_toning_o_hist}).
 
 The emittance function has three components, i.e. ${\bf e} = (e_r,e_g,e_b)$. Therefore the emitted light $I$  as a function of wavelength $\lambda$ for a given pixel $(x,y)$ on the electronic display is given by 
\begin{equation}
I(x,y,\lambda) =  \rho_r e_r(\lambda,{\bf \theta}) +\rho_g e_g(\lambda,{\bf \theta})  +\rho_b e_b(\lambda,\bf{\theta} ), 
\end{equation}
or
\begin{equation}
I(x,y,\lambda) =  {\bf \rho} \cdot {\bf e}(\lambda,{\bf \theta}).
\label{eq:emitted}
\end{equation}
Now consider the intensity of the light received by one pixel element at the camera sensor. Let $s_r(\lambda)$ denote the camera sensitivity function for the red component, then the red pixel value $I_c^{r}$ can be expressed as 
\begin{equation}
I_c^{r} \propto \int_{\lambda} \left [ {\bf \rho} \cdot {\bf e}(\lambda,{\bf \theta})) \right ] s_r(\lambda) d\lambda.
\label{eq:redpixel}
\end{equation}
Notice that the sensitivity function of the camera has a dependence on wavelength that is likely different than the emittance function of the monitor.  That is, the interpretation of ``red'' in the monitor is different from that of the camera.  Notice that a sign of proportionality is used in Equation~\ref{eq:redpixel} to specify that the pixel value is a linear function of the intensity at the sensor, assuming a linear camera and display. This assumption will be removed in  Section~\ref{sec:mainalgorithm}. 

Equation~\ref{eq:redpixel} can be written to consider all color components in the captured image  ${\bf{I_c}}$ as
\begin{equation}
{\bf I_c}  \propto  \int_{\lambda} \left [ {\bf \rho} \cdot {\bf e}(\lambda,{\bf \theta}) \right ] {\bf s}(\lambda) d\lambda.
\label{eq:camerapixel}
\end{equation}
where ${\bf s} = (s_r,s_g,s_b)$. 

\subsection{Message Structure}
% additive to enable convex optimization
 The pixel value ${\bf \rho}$ is controllable by the electronic display driver, and so it provides a mechanism for embedding information. 
 We use two sequential frames in our approach. We modify the monitor intensity by adding the value $\kappa$ and transmit two consecutive images, one with the added value ${\bf I_e}$ and one image of original intensity ${\bf I_o}$.  The recovered message depends on the display emittance function and camera sensitivity function if the embedded message is done by adding $\kappa$ as follows:
\vspace{-2pt}
\begin{equation}
{\bf I_e}  \propto  \int_{\lambda}\left [ ( \kappa +  {\bf \rho}) \cdot {\bf e}(\lambda,{\bf \theta}) \right ] {\bf s}(\lambda) d\lambda.
\label{eq:camerapixel1}
\end{equation}
Recovery of the embedded signal leads to a difference equation
\begin{equation}
{\bf I_e} -  {\bf I_o}  \propto  \int_{\lambda}\left [ ( \kappa) \cdot \bf{e}(\lambda,\bf{\theta})  \right ] {\bf s}(\lambda) d\lambda.
\label{eq:additive}
\end{equation}
The dependence on the properties of the display $e$ and the spectral sensitivity  of the camera $s$ remains.  
We use additive-based messaging, instead of ratio-based methods, because this structure is convenient for convexity of the algorithm as described in Section~\ref{sec:mainalgorithm}. 

\begin{figure*}[]
	\begin{center}
			\includegraphics[width=6.5in]{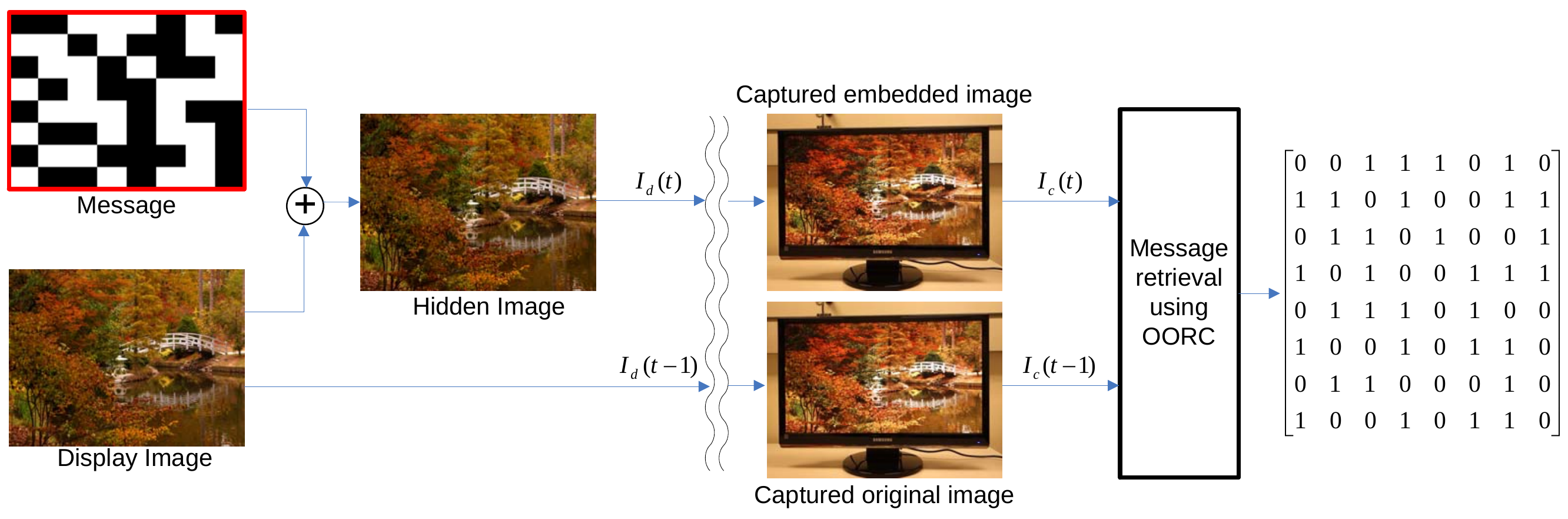}
				\end{center}
	\caption{Message Embedding and Retrieval.  Two sequential frames are sent, an original frame and a frame with an embedded message image.
	Simple differencing is not sufficient for message retrieval. Our method (OORC) is used to recover messages accurately.   }
	\label{fig:flowchart}
\end{figure*}

The main concept for message embedding is illustrated in Figure~\ref{fig:flowchart}. 
In order to convey many ``bits'' per image, we divide the image region into a series of block components.
 Each block can convey a bit ``1'' or ``0''. The blocks corresponding to a ``1''  contain the added value $\kappa$ typically set to 10 gray levels, while the zero blocks have no additive component ($\kappa=0$). 
The message is recovered by sending the original frame followed by a frame with the embedded message and using the difference for message recovery.  The message can also be added to the coarser scales of a image pyramid decomposition \cite{Burt83}, in order to better  hide the message within the display image content. The display can be tracked with existing methods \cite{Yuan12a}. 
This message structure is decidedly very simple, so the methods presented here can be applied to many message coding schemes.

When accounting for the nonlinearity in the camera and display, we rewrite Equation~\ref{eq:camerapixel} to include the radiometric response function $f$, 
 \begin{equation}
{\bf I_c}  = f\left(  \int_{\lambda} \left [ {\bf \rho} \cdot {\bf e}(\lambda,{\bf \theta}) \right ] {\bf s}(\lambda) d\lambda \right).
\end{equation}
More concisely,
 \begin{equation}
{\bf I_c}  = f\left(  {\bf I_d} \right),
\end{equation}
and the recovered display intensity is 
 \begin{equation}
{\bf I_d}  = f^{-1}\left(  {\bf I_d} \right) =g\left(  {\bf I_d} \right). 
\end{equation}
We use  polynomials to represent the radiometric inverse function $g(i)$.  The same inverse function $g$ is used for all color channels and gray-scale ratex patches.  This simplification of the color problem is justified by the accuracy of the empirical results. 

\subsection{Simultaneous Radiometric Calibration and Message Recovery via Convex Optimization}
\label{sec:mainalgorithm}
The two goals of message recovery and calibration can be combined to a single problem. 
While ideal  radiometric calibration would provide a  captured image that is a linear function of the displayed image, we show that calibrating followed by message recovery only gives a relatively small increase in message accuracy. 
However, if the two goals are combined into a simultaneous problem we have two benefits: 1) the problem formulation can be done in a convex optimization paradigm with a single global solution and 2) the accuracy increases significantly.

% Display the ideal and non-separable case
%\begin{figure}[!ht]
%  \centering
%  \subfigure[The displayed pair]{\includegraphics[width=0.23\textwidth]{imgs/img_ideal.pdf}}
%  \subfigure[The captured pair]{\includegraphics[width=0.23\textwidth]{imgs/img_actual.pdf}}
%  \caption{The feature space $(i_o,i_e)$ plot of the displayed training data (left) and captured training data (right).  After image capture, the two classes (one-bits and zero-bits) are not separable by a simple differencing of the original and embedding image.  In fact, there is no linear decision boundary that will separate the two classes.  The training data sequence is contained in the ratex patches.  A few outlier pixels near the borders are present but can be easily removed.}
%  \label{fig:separable}       % Give a unique label
%\end{figure}

Let $g(i)$ be the inverse function that is modeled with a fourth order polynomial as follows
\begin{equation}
g(i) = a_4i^4 + a_3i^3 + a_2i^2 + a_1i + a_0 .
\end{equation}
Consider two images frames $i_o$, where $i_o$ is the original frame and $i_e$ the image frame with the embedded message.  Note the use of  $i_o$ instead of ${\bf I_o}$ for notional compactness.  Since we are using an additive message embedding, we wish to classify the message bits as either ones or zeros based on the difference image $i_o - i_e$. 
In order to classify the message bits, the ratex patches are also used for training. Consecutive frames of ratex patches toggle between message bit ``1''  ($\kappa=10$) and message bit ``0'' ($\kappa = 0$). This training data can be used for a support vector machine (SVM) classifer. 
%However, consider Figure~\ref{fig:separable} which is a plot of typical training data in the $i_o,i_e$ feature space. The transmitted values at the display are separable, but the captured values are not linearly separable.

 Taking into account the radiometric calibration, we want to classify on the recovered data $g(i_o) - g(i_e)$.  
 Assuming that the inverse function can be modeled by a fourth order polynomial, the function to be classified is
\begin{equation}
\begin{array}{c}
g(i_o) - g(i_e)  =\\ a_4(i_o^4-i_e^4)  + a_3(i_o^3-i_e^3) + a_2(i_o^2-i_e^2) + a1(i_o-i_e).
\label{eq:higher}
\end{array}
\end{equation}
In Equation~\ref{eq:higher}, we see that the calibration problem has a physically motivated nonlinear mapping function. That is, we see that the original data ($i_o,i_e$) can be placed into a higher dimensional space using the nonlinear mapping function $\Phi$ which maps from a two dimensional space to a four dimensional space as follows
\begin{equation}
\Phi(i_o,i_e) =  \left[ \begin{array}{cccc} (i_o^4-i_e^4) & (i_o^3-i_e^3) &(i_o^2-i_e^2)& (i_o-i_e) \end{array} \right].
\end{equation}
In this four dimensional space we seek a separating hyperplane between the two classes (one-bits and zero-bits). Our experimental results indicate that these are not separable in lower dimensional space, but the movement to a higher dimensional space enables the separation.  Also, the form of that higher dimensional space is physically motivated by the need for radiometric calibration.
Therefore our problem becomes a support vector machine classifier where the support vector weights and the calibration parameters are {\it simultaneously} estimated.
That is, we estimate
\begin{equation}
w^Tu+b,
\end{equation}
where, $w\in \mathbf{R}^4$, $b$ are the separating hyperplane parameters, and $u$ is the input feature vector.  Since we want to perform radiometric calibration, the four-dimensional input is given 
\begin{equation}
u =  \left[ \begin{array}{cccc} a_4(i_o^4-i_e^4) & a_3(i_o^3-i_e^3) & a_2(i_o^2-i_e^2)& a(i_o-i_e) \end{array} \right]^T.
\end{equation}
Notice that the $w^Tu+b$ is still linear in the coefficients of the inverse radiometric function. These coefficients and the scale factor $w$ are estimated simultaneously.  We arrive at the important observation that accounting for the CDTF preserves the convexity of the overall classification problem.  The coefficients of the function $g$ are scaled by $w$, so  that calibration and classification can be done {\it simultaneously}, and convexity of the SVM is preserved. 

\subsection{Radiometric Calibration with Hidden Ratex}
\label{sec:histeq}
The main disadvantage of OORC is the requirement that visible ratex patches must be placed on screen. Ratex patches are somewhat visually obtrusive and unattractive for certain applications. However, they are convenient for modeling the CDTF.
Instead of directly observing the effects of the CDTF on the full intensity gamut, we can observe how the CDTF modifies the intensity histogram. 
For this to work, we need to know the initial intensity distribution of an image before it passes through the CDTF.
We perform an intensity mapping on every image entering the camera-display transfer function so the intensity histogram is known.
We can think of the known intensity mapping of these images as ``hidden ratex.''
Once the image is camera-captured, the new, modified distribution of the image's intensities are observed.
Since the intensity distribution is predetermined, we are able to measure the effects of the CDTF by observing the differences in the camera-captured intensity histogram.
For example, we may wish to choose a uniform, or near uniform intensity distribution for camera-display transfer images.
By histogram equalizing a displayed image, a receiver can infer that the distribution of this image's intensities are near uniform.
An intensity mapping is applied to an image before it is displayed.
Although this will have an effect on the appearance of the carrier image, we refer to this method as hidden ratex because it does not require markers to be displayed on screen for calibration.
Once the image is captured, the photometric effects of the CDTF has altered the image.
The captured image is then intensity mapped again, so that its intensity histogram is more similar to the displayed distribution, before distortion by the CDTF.
In other words, histogram intensity mapping acts as the inverse CDTF.
Although there is not one-to-one correspondence, intensity mapping is an effective method for hidden ratex as a visibly non-disruptive method for radiometric calibration.

Because histogram-driven intensity mapping serves as an effective inverse-CDTF mapping, embedded messages bits can be labeled with simple thresholding.
For each pair of corrected images (original and embedded), intensity mapping is applied to the original image. 
That same mapping is then applied to the embedded message.
The difference between the original and carrier image are then computed.
The embedded blocks are now separable by a simple constant threshold, because, undisrupted by the photometric effects of the CDTF, message blocks are nothing more than a known added constant.
In other words, $i_e$ and $i_o$ are remapped via the same intensity mapping.
The remapped difference $i_e-i_o$ is used to recover the message bit.

%%%%%%%%%%%%RESULTS%%%%%%%%%%%%%%%%%%%%%%

\section{Results}

For empirical validation, 9 different combination of displays and cameras are used comprised of 3 displays: 1) LG 
M3204CCBA 32 inch, 2) Samsung SyncMaster 2494SW,  3) iMac (21.5 inch 2009); and 3 cameras: 1) Canon EOS Rebel XSi, 2) Nikon D70, 3) Sony DSC-RX100.
Fifteen 8-bit display images are used. % illustrated in Figure~\ref{fig:testImgs}. 
From each display image, we create a display video of 10 frames: 5 frames with the original display images interleaved with 5 images of embedded time-varying messages.  An embedded message frame is followed by an original image frame to provide the temporal image pair $i_e$ and $i_o$.   The display image does not change in the video, only the bits of the message frames.   
Each message frame has $8\times 8=64$ blocks used for message bits (with 5 bits used for ratex patches for calibration and classification training data).  
Considering 5 display images, with 5 message frames and 59 bits per frame results in approximately 1500 message bits. 
The accuracy for each video is defined as the number of correctly classified bits divided by the total bits embedded and is averaged over all testing videos.  The entire test set over all display-camera combinations is approximately 18,000 test bits.

There are 4 methods for embedded message recovery.
 Method 1 has no radiometric calibration,  only the difference $i_e-i_o$ is used to recover the message bit.   
 Method 2 is calibration followed by differencing for message recovery.  
 Method 3  (OORC) is the optimal calibration where both radiometric calibration and classification are done simultaneously.
 Method 4 is calibration via hidden ratex followed by simple differencing for message recovery. 
 For the first three methods, training data from pixels in the ratex patches are used to train an SVM classifier. 
 For each of the 9 display-camera combinations, the accuracy of the 4 message recovery methods was tested with 2 sets of experimental variables: 
 1) 0\degree frontal camera-display view;
 2) 45\degree oblique camera-display view;
and:
1) embedded message intensity difference of 5;
2) embedded message intensity difference of 3.
The results of these tests are can be found in Tables~\ref{tab:oblique3},~\ref{tab:front3},~\ref{tab:oblique5}, and~\ref{tab:front5}.

\begin{table}[h]
	\centering
		\begin{tabular}{|p{1.5cm}|p{1.5cm}|p{1cm}|p{1cm}|p{1cm}|} \hline
			Accuracy (\%)	& Naive Threshold	& Two-step	& OORC	& Hidden Ratex \\ \hline
			Canon-iMac		& 72.94		& 75.67	& 99.17	& 89.63 \\ \hline
			Canon-LG		& 58.94		& 84.94 	& 98.44	& 95.74 \\ \hline
			Canon-Samsung 	& 48.44 		& 64.89 	& 99.39 	& 89.91 \\ \hline
			Nikon-iMac 		& 60.17 		& 75.50 	& 95.17 	& 90.00 \\ \hline
			Nikon-LG 		& 49.72 		& 73.39 	& 99.33 	& 94.81 \\ \hline
			Nikon-Samsung 	& 47.22 		& 72.89 	& 95.00 	& 89.54 \\ \hline
			Sony-iMac 		& 64.44 		& 76.00 	& 99.06 	& 71.11 \\ \hline
			Sony-LG		&  56.11 		& 75.61 	& 98.56 	& 90.93 \\ \hline
			Sony-Samsung 	& 47.50 		& 79.11 	& 98.89 	& 87.80 \\ \hline
			Average 		& 56.17 		& 75.33 	& 98.11 	& 88.83 \\ \hline
		\end{tabular}
	\vspace{5pt}
	\caption{ Accuracy of embedded message recovery and labeling with additive difference +3 on [0,255] and captured with 45\degree oblique perspective. }
	\label{tab:oblique3}
\end{table}

\begin{table}[h]
	\centering
		\begin{tabular}{|p{1.5cm}|p{1.5cm}|p{1cm}|p{1cm}|p{1cm}|} \hline
			Accuracy (\%)	& Naive Threshold	& Two-step	& OORC	& Hidden Ratex \\ \hline
			Canon-iMac		& 85.56		& 83.06	& 96.44	& 91.57 \\ \hline
			Canon-LG		& 86.39		& 90.94 	& 98.67 	& 94.07 \\ \hline
			Canon-Samsung 	& 87.94 		& 87.78 	& 98.94 	& 91.30 \\ \hline
			Nikon-iMac 		& 84.06 		& 84.00	& 96.50 	& 90.27 \\ \hline
			Nikon-LG 		& 74.67 		& 81.44 	& 99.94 	& 90.09 \\ \hline
			Nikon-Samsung 	& 77.33 		& 86.06 	& 98.00 	& 91.57 \\ \hline
			Sony-iMac 		& 89.33 		& 84.22 	& 99.44 	& 70.00 \\ \hline
			Sony-LG		& 87.61 		& 95.39 	& 99.72 	& 80.74 \\ \hline
			Sony-Samsung 	& 80.00 		& 83.78 	& 96.26 	& 84.54 \\ \hline
			Average 		& 83.56		& 86.30 	& 98.22 	& 87.13 \\ \hline
		\end{tabular}
	\vspace{5pt}
	\caption{  Accuracy of embedded message recovery and labeling with additive difference +3 on [0,255] and captured at 0\degree frontal view. }
	\label{tab:front3}
\end{table}

\begin{table}[h]
	\centering
		\begin{tabular}{|p{1.5cm}|p{1.5cm}|p{1cm}|p{1cm}|p{1cm}|} \hline
			Accuracy (\%)	& Naive Threshold	& Two-step	& OORC	& Hidden Ratex \\ \hline
			Canon-iMac		& 97.06		& 94.50	& 99.83	& 95.37 \\ \hline
			Canon-LG		& 87.89		& 99.00 	& 99.39 	& 99.44 \\ \hline
			Canon-Samsung 	& 71.67		& 88.11 	& 100.00 	& 95.37 \\ \hline
			Nikon-iMac 		& 91.89		& 93.67 	& 96.00	& 96.11 \\ \hline
			Nikon-LG 		& 81.56		& 95.11 	& 99.94 	& 98.88 \\ \hline
			Nikon-Samsung 	& 58.78 		& 92.22 	& 99.39 	& 97.41 \\ \hline
			Sony-iMac 		& 92.28 		& 92.00 	& 99.72 	& 80.37 \\ \hline
			Sony-LG		& 77.06 		& 96.22 	& 100.00 	& 91.13 \\ \hline
			Sony-Samsung 	& 63.28 		& 94.17 	& 99.89 	& 81.67 \\ \hline
			Average 		& 80.16 		& 93.89 	& 99.35 	& 93.71 \\ \hline
		\end{tabular}
	\vspace{5pt}
	\caption{  Accuracy of embedded message recovery and labeling with additive difference +5 on [0,255] and captured with 45\degree oblique perspective. }
	\label{tab:oblique5}
\end{table}

\begin{table}[h]
	\centering
		\begin{tabular}{|p{1.5cm}|p{1.5cm}|p{1cm}|p{1cm}|p{1cm}|} \hline
			Accuracy (\%)	& Naive Threshold	& Two-step	& OORC	& Hidden Ratex \\ \hline
			Canon-iMac		& 95.28		& 96.61	& 99.00	& 95.74 \\ \hline
			Canon-LG		& 97.11		& 99.72 	& 97.17 	& 97.59 \\ \hline
			Canon-Samsung 	& 97.39 		& 97.33 	& 98.94 	& 94.35 \\ \hline
			Nikon-iMac 		& 98.39 		& 99.17 	& 99.22 	& 96.11 \\ \hline
			Nikon-LG 		& 99.83 		& 100.00 	& 99.83 	& 97.31 \\ \hline
			Nikon-Samsung 	& 96.33 		& 97.44 	& 98.56 	& 95.74 \\ \hline
			Sony-iMac 		& 97.72 		& 97.00 	& 99.94 	& 81.67 \\ \hline
			Sony-LG		& 99.39 		& 100.00 	& 100.00 	& 90.74 \\ \hline
			Sony-Samsung 	& 92.50 		& 92.33	& 98.06	& 90.28 \\ \hline
			Average 		& 97.10 		& 97.73 	& 98.97	& 93.28 \\ \hline
		\end{tabular}
	\vspace{5pt}
	\caption{  Accuracy of embedded message recovery and labeling with additive difference +5 on [0,255] and captured at 0\degree frontal view. }
	\label{tab:front5}
\end{table}

\section{Discussion and Conclusion}
\label{sec:discussion}

The results indicate a substantial improvement of bit classification in a camera-display messaging system with our methods. We demonstrate experimental results for nine different camera-display combinations at frontal and oblique viewing directions. We show that naive thresholding is a poor choice because the variation of display intensity with camera position is ignored. Any method that embeds a message without accounting for the variation of display intensity will degrade for non-frontal views. We present two ways to perform online radiometric calibration. The first method uses calibration information in the image called ratex patches. In the second approach, the calibration information is hidden and no patches appear in the image.
Our experimental results show that hidden, dynamic messages can be embedded in a display image and recovered robustly. We show that naive methods of message embedding without photometric modeling lead to failed message recovery, especially for oblique views (45\degree) and small intensity messages (+3).
We present a visually non-disruptive method for radiometric calibration in the form of hidden ratex intensity mapping.
Although the CDTF is spatially dependent, a single set of calibration coefficients per frame were sufficient for high message accuracy.
The approach is well-justified by theory and by empirical evaluation.

{
\bibliographystyle{ieee}
\bibliography{cvpr2015}
}

\end{document}